# Modified FOX Optimizer for Solving optimization problems

Dler O. Hasan [a,*], Hardi M. Mohammed [a] and Zrar Khalid Abdul [a]

[a] Department of Computer Science, College of Science, Charmo University, 46023 Chamchamal/Sulaimani, Kurdistan Region, Iraq.

\* dler.osman@chu.edu.iq

**Abstract:** The FOX optimizer, inspired by red fox hunting behavior, is a powerful algorithm for solving real-world and engineering problems. However, despite balancing exploration and exploitation, it can prematurely converge to local optima, as agent positions are updated solely based on the current best-known position, causing all agents to converge on one location. This study proposes the modified FOX optimizer (mFOX) to enhance exploration and balance exploration and exploitation in three steps. First, the Oppositional-Based Learning (OBL) strategy is used to improve the initial population. Second, control parameters are refined to achieve a better balance between exploration and exploitation. Third, a new update equation is introduced, allowing agents to adjust their positions relative to one another rather than relying solely on the best-known position. This approach improves exploration efficiency without adding complexity. The mFOX algorithm's performance is evaluated against 12 well-known algorithms on 23 classical benchmark functions, 10 CEC2019 functions, and 12 CEC2022 functions. It outperforms competitors in 74% of the classical benchmarks, 60% of the CEC2019 benchmarks, and 58% of the CEC2022 benchmarks. Additionally, mFOX effectively addresses four engineering problems. These results demonstrate mFOX's strong competitiveness in solving complex optimization tasks, including unimodal, constrained, and high-dimensional problems.

## 1. Introduction

Optimization is a crucial process for identifying the best possible solution from a wide range of alternatives to achieve desired outcomes, such as maximizing or minimizing specific objectives [1]. With the growing complexity of real-world problems, traditional optimization methods like gradient descent and linear programming often face significant challenges [2]. These deterministic techniques are prone to trap into local optima, especially when they applied to high-dimensional or multimodal problems, where search spaces are vast and complex. As a result, deterministic approaches struggle to provide reliable solutions for many modern applications. To overcome these limitations, field of optimization has advanced with the development of metaheuristic algorithms. Unlike traditional methods, metaheuristics utilize a stochastic approach that focuses on balancing exploration, by searching across different areas of the problem space, and exploitation, by refining solutions within promising regions [3]. This enables them to better navigate complex, multidimensional landscapes and avoid local optima. These algorithms draw inspiration from diverse sources such as natural phenomena [4], biological principles [5], animal behaviors [6], and human strategies [7]. Based on these sources, researchers have proposed a wide range of metaheuristic algorithms to achieve more effective solutions. These algorithms are typically grouped into four main categories based on their source of inspiration: evolutionary-based, swarm-based, physics/chemistry-based, and human-based, as shown in Figure 1. As a result, they are capable of solving a wide array of complex problems, including network design, vehicle routing, supply chain management, and machine learning hyperparameter optimization [8].

The class of algorithms known as evolutionary-based algorithms is based on the principles of natural selection, where candidate solutions are gradually improved through selection, crossover, and mutation processes. These algorithms are very useful for finding solutions to difficult optimization problems by exploring large solution spaces and fine-tuning for the best results. Some of the most known are the Genetic Algorithm (GA) [9] and the Differential Evolution (DE) [10], which contain such operators as crossover and mutation based on the Darwinian evolution. Also, Evolutionary Strategies (ES) [11] uses principles of biological evolution to come up with the best solutions.

Swarm-based algorithms mimic the natural behaviors of swarms as seen in the natural world, for instance, foraging and hunting behavior of animals. These methods mimic interactions between individuals within a swarm or group so as to search and optimize solution spaces effectively [12]. Popular examples include

Particle Swarm Optimization (PSO) [13] that is based on bird and fish social foraging, and Ant Colony Optimization (ACO) [14], which models the pathfinding abilities of ants. Other algorithms include Grey Wolf Optimization (GWO) [15] which is based on the hunting behavior of grey wolves; Marine Predator Algorithm (MPA) [16] which is based on predator-prey behavior in marine environment. TSA (Tunicate Swarm Algorithm) [17], GJO (Golden Jackal Optimization) [18] are also included in this category based on the behaviors of tunicates and golden jackal. Whale Optimization Algorithm (WOA) [19] mimics the behavior of humpback whales, and its design is based on the bubble-net hunting strategy. ChOA (Chimpanzee Optimization Algorithm) [20] is based on hunting behavior of chimpanzees as a social group that is based on intelligence and sexual selection. The algorithm is based on the decision-making of chimpanzees, unlike other social predators that have different mechanisms during group hunting. Furthermore, Fitness Dependent Optimizer (FDO) [21] is based on the reproductive phase of bees and especially focuses on the decision-making process of bees. Dragonfly Algorithm (DA) [22] draws its primary inspiration from the natural swarming behaviors of dragonflies, specifically their static and dynamic modes of group movement. American Zebra Optimization Algorithm (AZOA) [23] has been derived from the unique social structure and leadership pattern of the American zebra herds. Hippopotamus Optimization (HO) [24] is an algorithm based on the behavior of hippos, and includes a three-phase model of position in water, defense against predators, and escape mechanisms. Remora optimization algorithm (ROA) [25] is mainly based on the parasitic behavior of remoras. Newton-Raphson-Based Optimizer (NRBO) [26] leverages Newton-Raphson's method, utilizing two key mechanisms: the Newton-Raphson Search Rule and Trap Avoidance Operator, as well as particular matrix groups for the purpose of improving the result exploration. The Walrus Optimizer (WO) [27] takes its inspiration from the natural behaviors of walruses, including their migration patterns, breeding practices, social gatherings, feeding habits, and responses to environmental threats and safety cues.

Human-based algorithms mimic human activities and social interactions in order to solve optimization problems. Some of these are the Driving Training-Based Optimization (DTBO) [28] which is based on the learning between a driving trainer and trainee and the Technical and Vocational Education and Training-Based Optimizer (TVETBO) [29] which is based on learning in technical training situations. Also, Doctor and Patient Optimization (DPO) [30] is the optimization model of the relationship between physicians and patients, and Hiking Optimization Algorithm (HOA) [31] takes inspiration from group hiking behavior.

Physics- and chemistry-based algorithms use principles of the physical and chemical sciences to enhance the optimization of problem-solving procedures. For example, Kepler Optimization Algorithm (KOA) [32] is based on Kepler's laws of planetary motion and Gravitational Search Algorithm (GSA) [33] is based on Newton's law of gravity. The Water Cycle Algorithm (WCA) [34] mimics the river formation and the Equilibrium Optimizer (EO) [35] uses mass balance models to find the best states. Also, Spring Search Algorithm (SSA) [36] is based on the Hooke's law of elasticity. These categories show that metaheuristic algorithms may be inspired by a wide variety of motivations, all of which are tailored to address different kinds of optimization problems.

Metaheuristic algorithms are stochastic methods that look for good feasible solution within the solution space in order to solve optimization problems. The nature of the optimization problems as unimodal and multimodal makes it compulsory for metaheuristic algorithms to strike a balance between exploitation and exploration. The FOX optimizer which was recently developed [37] is well known for its effectiveness in solving optimization problems. However, it has a limitation in the search space exploration, and gets stuck in local optima most of the times as the agents are mostly moving around the best-known position. In this paper, a modified version of the FOX algorithm is proposed through three steps. First, the OBL strategy is applied to improve the initial population. Second, the exploration phase is enhanced by adjusting the control parameter ($a$) and removing the $MinT$ variable, as these modifications have a significant impact on improving the exploration phase and achieving a better balance between exploration and exploitation. Third, to further enhance exploration, a new equation is introduced in which the positions of the foxes are updated based on each other, rather than solely on the best-known position. This modification aims to broaden the exploration of the search space. Most importantly, this new approach is implemented without increasing the complexity of the proposed mFOX algorithm compared to the original FOX optimizer. Moreover, these adjustments allow the agents (foxes) to explore the entire search space more effectively, thereby avoiding entrapment in local optima. The key contributions of this study are as follows:

- A novel technique called the modified Fox-inspired Optimization Algorithm (mFOX) has been developed.
- The performance of mFOX is evaluated on 23 classical benchmark functions, 10 CEC2019 benchmarks, 12 CEC2022 benchmarks, and 4 real-world engineering problems.
- The experimental results, validated by the Wilcoxon rank-sum statistical test.
- The performance of mFOX is assessed against twelve well-known metaheuristic algorithms, including WOA, TSA, ChOA, FDO, GWO, DA, AZOA, HO, ROA, WO, NRBO, and FOX, using statistical data analysis, convergence analysis, runtime analysis, and the Tied Rank (TR) method.

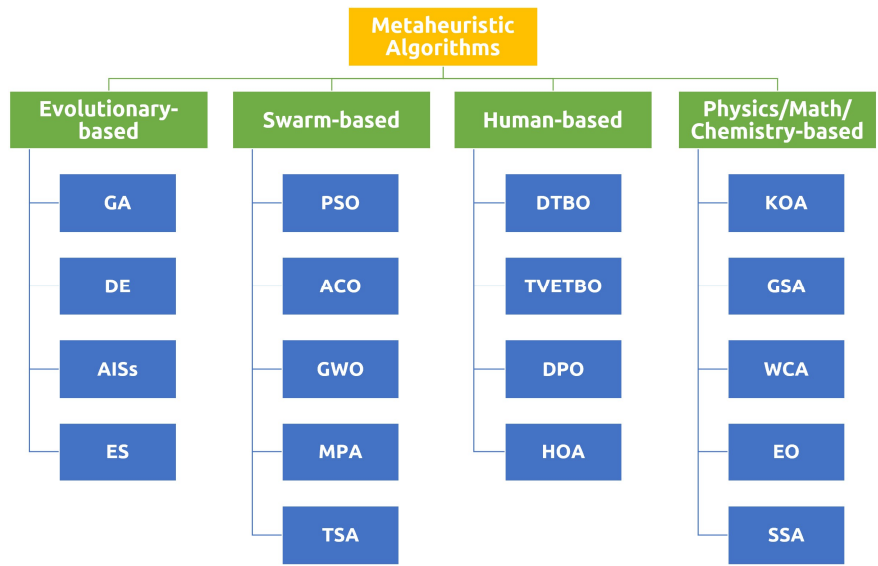

**Figure 1.** Nature-inspired metaheuristic algorithms.

## 2. Methodology

This section presents descriptions of both the original FOX algorithm and the newly proposed mFOX algorithm.

### 2.1. The initial version of the FOX algorithm

Nature has developed highly effective and efficient behaviors to solve complex problems in a wide range of situations, such as foraging, hunting, migration, shelter building, territory defense, mate selection, and cooperation in social structures. Foraging is a process of searching for food in which animals have to decide between the exploration of new territories and the utilization of resources that have been found before; hunting is another kind of activity in which animals have to use the group tactics in order to achieve the desired result. The migration reveals the efficient ways of energy spending for navigation and resource utilization; nest building or shelter construction illustrates the proper utilization of available resources. Cooperation is seen in social animals in activities such as defense of territory and protection of young, as well as in searching for food and other resources, where social animals work together to increase their chances of survival in a world that is unpredictable. Such natural behaviors inspire the development of metaheuristic algorithms, which are valued for their high accuracy, quick convergence to optimal solutions, and minimal computational demands, making them ideal for tackling complex optimization problems.

Recently, a new algorithm, named FOX optimizer, was introduced by [37], inspired by the hunting techniques of red foxes, particularly their unique method of diving into snow to catch prey. This diving technique represents the exploitation phase of the algorithm. When the snow obscures the fox's view, it employs a random walk in the search area as part of its exploration phase to locate the prey. Therefore, the FOX algorithm incorporates these two behaviors—random walking and jumping—to effectively solve optimization problems. At first, the red fox navigates randomly through the search space to locate its prey, relying on its ability to hear the ultrasonic sounds emitted by the prey. This random movement serves as the inspiration for the exploratory behavior implemented in FOX. While foraging, the fox may detect these sounds, marking its transition into the exploitation phase. Since the sound takes time to reach the fox, the distance it travels can be calculated by multiplying the time by the speed of sound in air, which is approximately 343 meters per second. However, due to the static nature of this value, an alternative approach is employed to assess sound propagation in air. As the fox advances toward the prey, it prepares to jump based on the time it takes for the sound to reach it. Researches [38], [39] indicate that foxes tend to prefer jumping in a northeastern direction due to magnetic alignment, resulting in an 82% success rate for catching prey in that direction. Conversely, if they jump in the opposite direction, the success rate drops to 18%. Thus, it can be concluded that there are two primary strategies for capturing prey.

### 2.1.1. FOX optimizer: mathematical model

In FOX optimization, the initialization process begins by generating a set of random solutions to form the population. This FOX population matrix is established using Eq.(1), which represents the location of each red fox. Here, *pop* refers to the number of potential solutions, *dim* denotes the dimensionality of the problem, and $X_{ij}$ represents the position of the $i^{th}$ fox in the $j^{th}$ dimension. This method ensures the distribution of each fox's position within the search space.

$$X = \begin{bmatrix} X_{1.1} & X_{1.2} & \cdots & \cdots & X_{1.dim} \\ X_{2.1} & X_{2.2} & \cdots & \cdots & X_{2.dim} \\ \vdots & \vdots & \vdots & \vdots & \vdots \\ X_{pop.1} & X_{pop.2} & \cdots & \cdots & X_{pop.dim} \end{bmatrix} \quad (1)$$

In each iteration, the fitness of every search agent is evaluated. After comparing the fitness values of the agents' positions, the position with the highest fitness ($BestX$) and its corresponding fitness value ($BestFitness$) are identified as the prey's location. A random variable, $r$, is introduced to evenly balance the exploration and exploitation phases within each iteration. In FOX, this variable assigns a 50% probability to either exploration or exploitation, meaning roughly half of the iterations focus on exploration while the other half are dedicated to exploitation. This approach is crucial for maintaining balance and preventing the algorithm from getting stuck in local optima. To ensure this balance, a conditional statement is employed to divide the iterations equally between exploration and exploitation.

Additionally, another variable, $a$, is introduced to gradually reduce the search performance based on the best solution, $BestX$. With each iteration, the value of $a$ decreases, leading to improved pursuit of the target, as the agent moves closer to the prey in every step. Alongside this, the fitness value influences the search agents by allowing them to bypass local optima. If the new position does not lead to a significant change, the exploration phase is temporarily paused, enabling other phases to become active.

### 2.1.2. Exploitation Phase

The jumping behavior of the red fox illustrates its exploitation phase, characterized by an 82% probability of successfully capturing prey when it jumps in the northeastern direction, compared to only an 18% chance when jumping in the opposite direction. The random variable $p$ ranges from 0 to 1. Thus, if the generated random number $p$ is either above or below 0.18, a new position for the red fox must be determined. To establish this new position, the sound travel distance $DistST_{it}$ needs to be calculated, along with the distance from the red fox to the prey $DistFoxToPrey_{it}$, and the jumping valu $Jump_{it}$. To calculate the distance of the sound originating from the red fox, the speed of sound in air $SpS$ (v*elocity*) is multiplied by the sound travel time $TimeST_{it}$ (*time*), as expressed in Eq. (2). The value of $TimeST_{it}$ is a random number within the range [0, 1].

$$DistST_{it} = SpS \times TimeST_{it} \quad (2)$$

To calculate $SpS$, as indicated in eq. (3), the best position found so far $BestPosition_{it}$ is divided by the time it takes for sound to travel from the fox to the prey $TimeST_{it}$.

$$SpS = \frac{BestPosition_{it}}{TimeST_{it}} \quad (3)$$

After calculating the sound travel distance, the distance between the fox and the prey $DistFoxPrey_{it}$ is determined by dividing the sound travel distance by 2 (or equivalently, multiplying by 0.5), as per the principles of physics, where the sound wave must travel to the object and then return to the detector. This is demonstrated in Eq. (4).

$$DistFoxPrey_{it} = DistST_{it} \times 0.5 \quad (4)$$

After determining the distance between the fox and its prey, the fox must calculate the jump height $Jump_{it}$, which is done using Eq. (5). Here, 9.81 represents the acceleration due to gravity, and $t$ corresponds to the average time it takes for sound to travel. The time is squared to account for both the upward and downward phases of the jump. The time transition $tt$ value is calculated by dividing the sum of $TimeST_{it}$ across all dimensions, as shown in Eq. (6). The average time $t$ is then computed by dividing the transition time $tt$ by 2. This reflects the fact that the jump is divided into two distinct phases: ascent and descent. To account for this, both the average time and gravitational force are multiplied by 0.5, representing the separate time intervals for upward and downward motion. Therefore, gravity and average time are both scaled by 0.5.

$$Jump_{it} = 0.5 \times 9.81 \times t^2 \quad (5)$$

$$tt = \frac{\sum TimeST_{it}(i,:)}{dimension} \quad (6)$$

Finally, the new position of the fox is determined based on the value of $p$, as shown in Eq. (7). If $p$ is greater than 0.18, the $Jump$ value is multiplied by $DistFoxPrey_{it}$, and $c_1$, where $c_1$ ranges from [0, 0.18] (in this case, $c_1 = 0.18$). This calculation applies when the red fox jumps in the northeastern direction. Conversely, if $p$ is less than 0.18, the $Jump$ value is multiplied by $DistFoxPrey_{it}$ and $c_2$, where $c_2$ ranges from [0.19, 1] (here, $c_2 = 0.82$), which occurs when the fox jumps in the opposite direction.

$$X_{(it+1)} = \begin{cases} DistFoxPrey_{it} \times Jump_{it} \times c_1, & p > 0.18 \\ DistFoxPrey_{it} \times Jump_{it} \times c_2, & p \leq 0.18 \end{cases} \quad (7)$$

The parameters $c_1$ and $c_2$ are set at 0.18 and 0.82, respectively, reflecting the jumping patterns of a red fox, which may either jump northeast or in the opposite direction. These values primarily determine the strength of the exploitation phase. If the $p$ value exceeds 0.18, it indicates that the red fox will jump toward the northeast. To determine a new position in this case, both $DistFoxPrey_{it}$ and $Jump_{it}$ are multiplied by $c_1$, thereby increasing the likelihood of moving toward optimal solutions. Conversely, if the $p$ value is less than 0.18, indicating a low probability (18%) of capturing prey, the fox jumps in the opposite northeast direction. In this scenario, both $DistFoxPrey_{it}$ and $Jump_{it}$ are multiplied by $c_2$.

### 2.1.3. Exploration Phase

To regulate the random walk, the fox explores its environment based on its best-known position. To facilitate a random movement toward this optimal position, a minimum time variable $MinT$ and the variable $a$ are employed to govern the search process. Equations (8) and (9) detail the calculations for the $MinT$ and $a$ variables, with $MinT$ determined by identifying the minimum of $tt$. In each iteration, the value of $a$ is calculated using the current iteration and the maximum iteration, as shown in Eq. (9). This equation decreases the value of $a$ from 2 to 0, allowing the red foxes to get closer to their prey with each iteration.

$$MinT = Min(tt) \quad (8)$$

$$a = 2 \times \left(1 - \frac{it}{Max_{it}}\right) \tag{9}$$

Eq. (10) illustrates the exploration technique employed by the fox in its search for a new position in the search space $X_{(it+1)}$. The use of $rand(1, dimension)$ enables the fox to move stochastically, facilitating its search for prey. To enhance the search capabilities of FOX, both $MinT$ and $a$ variable is incorporated.

$$X_{(it+1)} = BestX_{it} + rand(1, dimension) \times MinT \times a \tag{10}$$

### 2.1.4. Modified FOX (mFOX)

Metaheuristic algorithms must be efficient in maintaining a balance between exploration (searching new areas) and exploitation (refining the best-known solutions) within the search space. The optimal balance between exploration and exploitation depends on the specific problem and the characteristics of the algorithm. If an algorithm explores too much, it may waste time on suboptimal solutions. Conversely, if it focuses too heavily on exploitation, it may become trapped in local optima, unable to find the global best solution.

In the FOX algorithm, the exploitation phase (defined by Eq.(7)) and the exploration phase (defined by Eq. (10)) both rely on updating the agents' (foxes) positions in relation to the current best-known position, known as BestX. This approach causes the agents to move primarily around the best position (which is also the prey), leading to strong exploitation capabilities but weak exploration [40], [41]. The agents tend to remain close to the prey, limiting their ability to explore other regions of the search space. However, the parameter $a$ (which linearly decreases from 2 to 0) in Eq. (10) plays a crucial role in controlling the exploration and exploitation behaviors of the algorithm during the search process. Since this parameter is multiplied by $MinT$ parameter (which decreases from 1, or a value slightly less than 1, toward 0), the foxes' positions tend to move very close to the current best position (prey). As a result, this enhances exploitation but limits the foxes' ability to broadly explore the search space. To address this imbalance and improve exploration, mFOX optimization algorithm proposed as follows:

1. The OBL optimization technique, introduced by Tizhoosh in 2005 [42], enhances the quality of initial population solutions by incorporating diversity through the creation of opposite solutions. [42] has shown that an opposite candidate solution has a higher probability of being nearer to the global optimum compared to a randomly selected solution. This method involves generating an opposite solution for each candidate, allowing the fitness function values to assess both solutions and identify the superior one. Many meta-heuristic algorithms effectively adopt this principle to boost convergence rates and improve overall solution quality [43], [44], [45], [46]. The OBL strategy is defined as follows: consider a solution $S$ that has $n$ parameters, with each parameter restricted to the range $[lb_i, ub_i]$. An opposing solution $\hat{S} = (\hat{S}_1, \hat{S}_3, \hat{S}_1, \ldots, \hat{S}_n)$ is derived based on Eq. (11):

$$\hat{S}_i = lb_i + ub_i - S_i \tag{11}$$

Here, $lb_i$ and $ub_i$, represent the lower and upper bounds for the $i$th dimension, respectively. During optimization, the current solution $S$ is replaced by its opposite $\hat{S}$ if the opposite has a better fitness value. For each iteration, the fitness of both $S$ and $\hat{S}$ is calculated, and the fitter solution is chosen. For example, if $fit_{\hat{S}_i} < fit_{S_i}$, then $S = \hat{S}$; otherwise, $S$ remains unchanged for the next iteration [47].

2. The existing exploitation phase is retained, as it already demonstrates robust performance in refining solutions. However, we modified the main equation of the exploitation phase so that it only updates the best position when the new position is better than the current best position, as outlined in Eq. (12) and Eq. (13).

$$X_{new} = \begin{cases} DistFoxPrey_{it} \times Jump_{it} \times c_1, & p > 0.18 \\ DistFoxPrey_{it} \times Jump_{it} \times c_2, & p \leq 0.18 \end{cases} \tag{12}$$

$$BestX = \begin{cases} X_{new}, & fit_{X_{new}} < BestFitness \\ BestX, & else \end{cases} \tag{13}$$

3. In the FOX optimizer, the random walk (as defined in Eq. (10)) was originally intended to serve as the exploration technique but actually causes the foxes to move too close to the prey. This is because, in the initial version, the new positions of the foxes are calculated as the prey position (BestX) plus the parameter $a$ multiplied by $MinT$. The product of $a$ and $MinT$ results in a very small value, causing the new positions of the foxes to be very close to the prey (BestX). This leads to increased exploitation but weakens exploration. To address this issue, the random walk (Eq. (10)) has been redesigned to enhance exploration during the early iterations and to focus more on exploitation near the prey towards the end of the iterations by removing the $MinT$ parameter and modifying the $a$ parameter.

The parameter $a$ is modified such that, instead of decreasing from 2 to 0, it now decreases from 1 to 0 over the iterations, as defined by Eq. (14). However, since $a$ in the original FOX optimizer is multiplied by $MinT$, it produces a smaller value over the iterations compared to the modified $a$ (shown in Eq. (14)). A smaller value makes the foxes more likely to move towards the global best position (BestX), focusing on refining the solution in the current region. Figure 2 illustrates the changes in the $a$ parameter in both the FOX and mFOX algorithms, as well as the result of $a$ in the original FOX optimizer multiplied by $MinT$ over the iterations.

$$a_{modified} = 1 - \frac{2 \times it}{Max_{it}} + \frac{it^2}{Max_{it}^2} \tag{14}$$

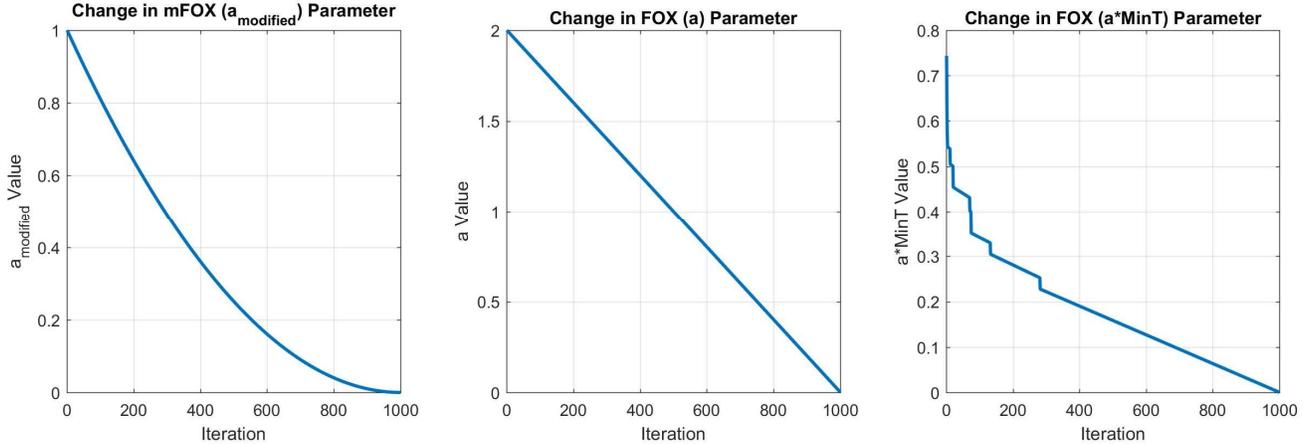

**Figure 2.** Change in mFOX($a_{modified}$), FOX (a) and FOX($a * MinT$) Parameters Over Iterations.

According to Figure 2, the result of the modified parameter $a$ in this paper is much larger compared to the parameter $a$ multiplied by $MinT$ over the iterations, which causes the new position of the foxes to be farther from BestX, especially at the beginning of the iterations. The random walk (Eq. (10)) is now modified as shown in Eq. (15). This equation updates the best position (BestX) only if the new position improves upon the current best, as demonstrated in Eq. (16).

$$X_{rwalk} = BestX_{it} + rand(1, dimension) \times a_{modified} \tag{15}$$

$$BestX = \begin{cases} X_{rwalk}, & fit_{X_{rwalk}} < BestFitness \\ BestX, & else \end{cases} \tag{16}$$

4. To further improve the exploration technique, a new equation, Eq. (17), is introduced to effectively adjust the positions of agents (foxes) throughout the search space. In Eq. (17), each fox's position is updated based on the positions of randomly selected foxes, rather than relying on the global best position (BestX). This method enhances the foxes' ability to thoroughly explore the search space. Additionally, the equation incorporates randomness into the foxes' movements, further boosting exploration capabilities. This modification expands the search radius, enabling the agents to explore more efficiently while retaining the FOX algorithm's strong exploitation strengths. As a result, the algorithm becomes more adept at avoiding local optima and finding better solutions. The variable $rN$ is randomly selected from the set $\{1, 1, 2\}$.

$$X_{(it+1)} = \begin{cases} X_{(it)} + \dfrac{rand(1, dim)}{rN} \cdot (X_{randSelFox} - rN \times X_{(it)}), & fit_{X_{randSelFox}} < fit_{X_{(it)}} \\ X_{(it)} + a_{modified} \cdot rand(1, dim) \cdot (X_{(it)} - rN \times X_{randSelFox}), & else \end{cases} \quad (17)$$

Figure 3 illustrates the hunting strategy employed by the mFOX algorithm. Furthermore, the specific details of the mFOX are presented in Algorithm 1 and Figure 4, which demonstrates the adjustments made to the original FOX optimization. As demonstrated in Algorithm 1, the balance between exploitation and exploration is maintained through the variable $r$, which ranges from 0 to 1. When $r$ is less than 0.2, the exploitation phase is executed. When $r$ is between 0.2 and 0.6, a random walk occurs in which the position of the foxes is changed based on the best solution (BestX), with the parameter $a_{modified}$ controlling the balance between exploration and exploitation. Conversely, when $r$ is between 0.6 and 1, the position of the foxes is changed based on other foxes, and again, the parameter $a_{modified}$ helps balance exploration and exploitation.

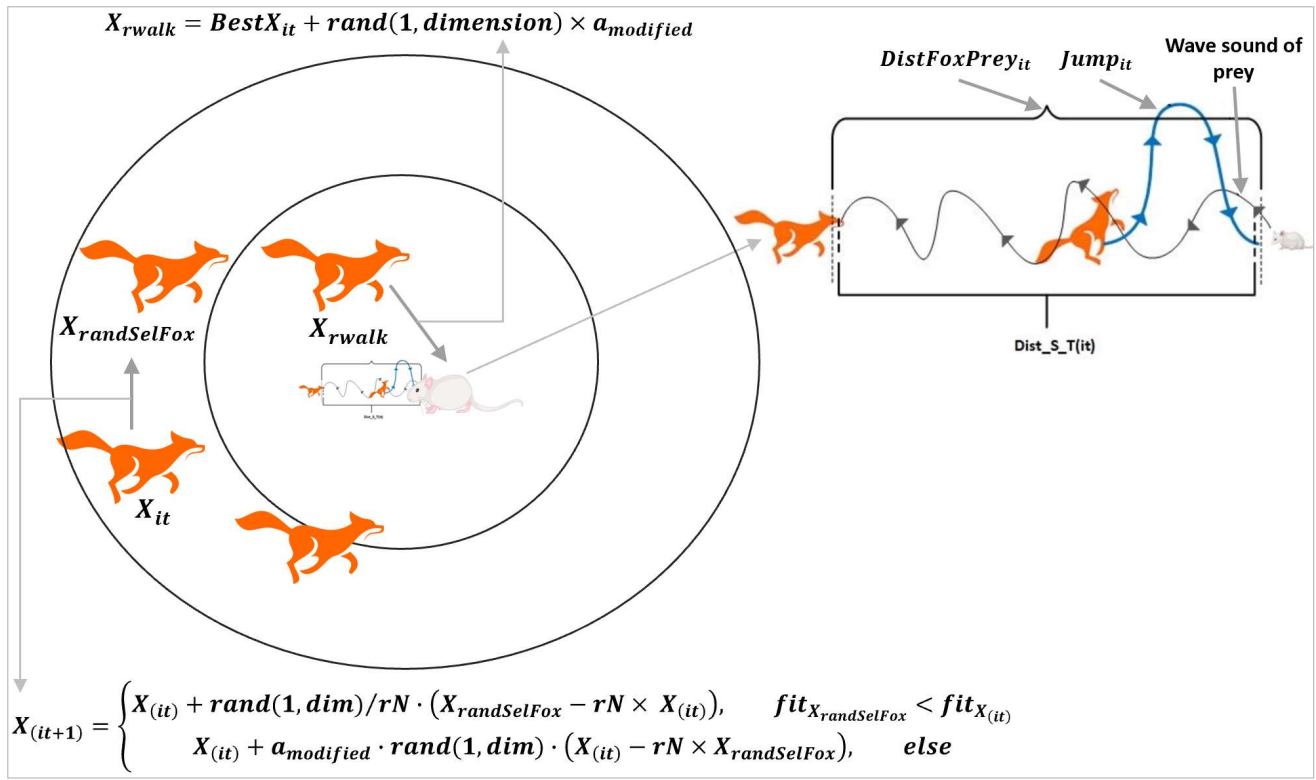

**Figure 3.** The hunting strategy of the mFOX [37].

| **Algorithm 1:** mFox Pseudocode |
|---|
| Initialize the red fox population **X** $i$ $(i=1, 2, ....., n)$ |
| Create opposite solutions using Eq. (11) |
| Evaluate the fitness of each solution and its opposite, then choose the fitter one. |
| **While** $it < Max_{it}$ |
| Initialize **$Dist\_S\_T$, $Sp\_S$, $Time\_S\_T$, $BestX$, $Dist\_Fox\_Prey$, $Jump$, $a_{modified}$, $BestFitness$.** |
| Calculate the fitness of each search agent |
| Select **$BestX$ and $BestFitness$ among the fox population** (X) in each iteration. |
|    If1 $fitness_i > fitness_{i+1}$ |
|      BestFitness=$fitness_{i+1}$ |
|      BestX=X(i, :) |

> **Endif1**
> Calculate $a_{modified}$ variable
> **If2** $r <= 0.2$
>     Initialize time randomly;
>     Calculate Distance_Sound_travels using Eq. (2)
>     Calculate $Sp\_S$ from Eq. (3)
>     Calculate distance from fox to prey using Eq. (4)
>     $tt$ = average time;
>     $t = tt/2$;
>     Calculate jump using Eq. (5)
>     **If3** $p > 0.18$
>         Find $X_{new}$ using Eq. (12) and Eq. (13)
>     **Elseif** $p <= 0.18$
>         Find $X_{new}$ using Eq. (12) and Eq. (13)
>     **EndIf3**
> **elseif2** $r < 0.6$
>     Explore $X_{rwalk}$ using Eq. (15) and Eq. (16)
> **elseif2** $r <= 1$
>     Explore $X_{(it+1)}$ using Eq. (17)
> **EndIf2**
> Check and amend the position if it goes beyond the limits
> Evaluate search agents by their fitness
> Update BestX
> $it = it + 1$
> **End while**
> return *BestX & BestFitness*

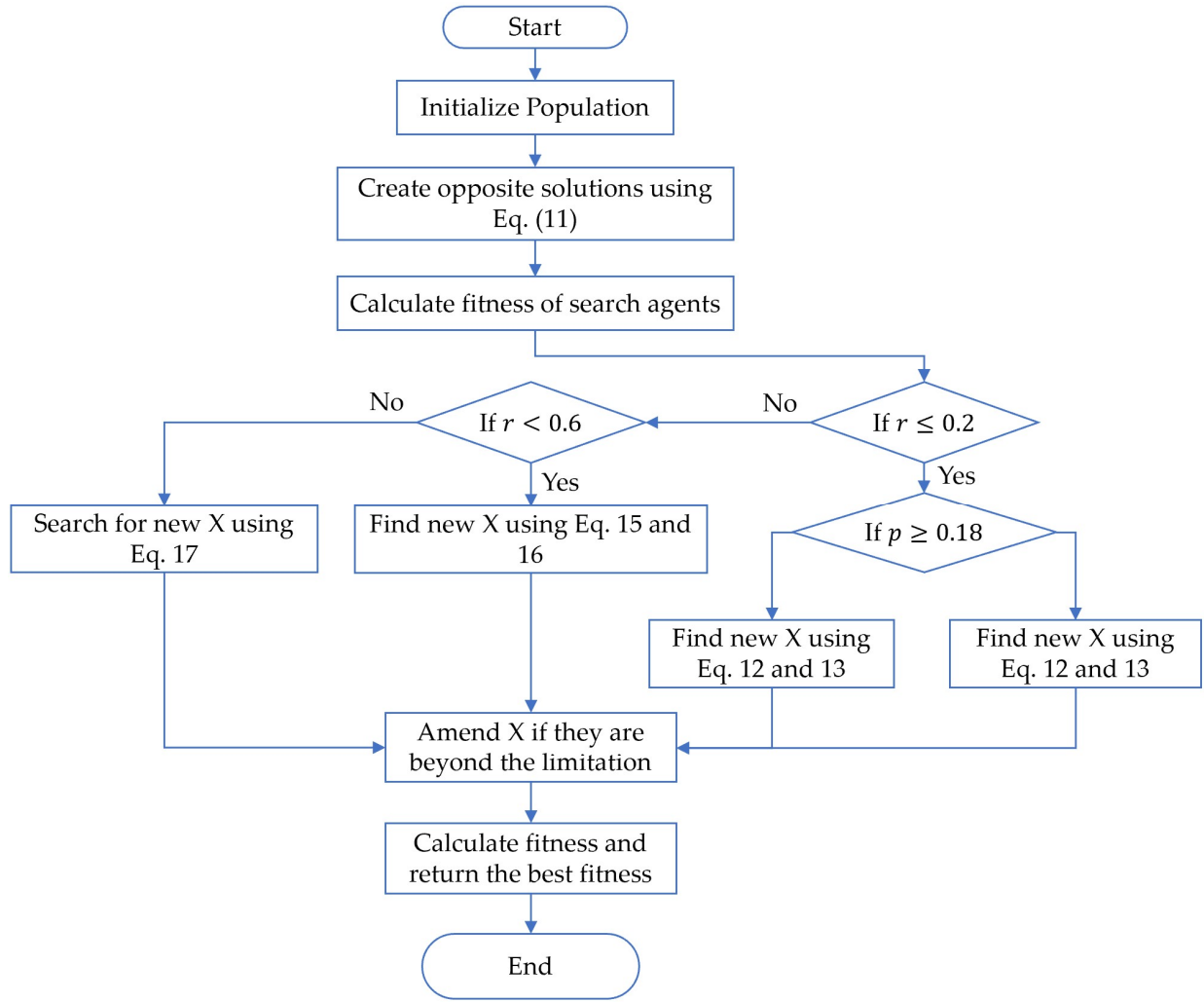

**Figure 4.** Flowchart of mFOX.

### 2.2. Algorithm complexity analysis

In this subsection, computational complexity of the mFOX algorithm is examined. The initialization phase, which includes constructing the population matrix and assessing the objective function, has a complexity of $O(Nm)$, where $N$ represents the number of foxes and m denotes the number of problem variables. In each iteration of mFOX, one objective function is evaluated, leading to a complexity of $O(NmT)$, with $T$ being the total number of iterations. Therefore, the overall computational complexity of the mFOX algorithm is $O(Nm(1 + T))$.

Table 1 presents the computational complexity of the competitor algorithms. The complexities are as follows: WOA, TSA, ChOA, GWO, DA, WO, NRBO, FOX, and mFOX all have a complexity of $O(Nm(1 + T))$. In comparison, AZOA is at $O(Nm(1 + 2T))$, while HO has a complexity of $O(Nm(1 + 5/2\,T))$. ROA's complexity is $O(Nm(1 + 3T))$, and FDO's complexity is $O(Nm(1 + 4T))$. These complexities are determined by the number of times the benchmark function is evaluated in each iteration. Therefore, to enable a fair comparison, the iteration count is adjusted for each metaheuristic algorithm during the simulation analysis to ensure that the total number of function evaluations (FEs) was uniform across all algorithms, as shown in Table 1.

**Table 1:** Parameters, Time Complexity, and Function Evaluations for mFOX and Comparison Algorithms.

| Algorithms | Year | Parameter | Value | Complexity | Max It | FEs |
|---|---|---|---|---|---|---|
| WOA | 2016 | Convergence parameter (a) | Linear reduction from 2 to 0. | $O(Nm(1+T))$ | 1000 | 30,000 |
| | | r | random vector in [0–1] | | | |
| | | l | random number in [−1, 1] | | | |
| TSA | 2020 | Pmin and Pmax | 1, 4 | $O(Nm(1+T))$ | 1000 | 30,000 |
| | | c1, c2, c3 | Random numbers in [0, 1] | | | |
| ChOA | 2020 | r1, r2, r3, u | ∈ [0, 1], ∈ [0, 1], ∈ [0, 2.5], ∈ [0,1], | $O(Nm(1+T))$ | 1000 | 30,000 |
| FDO | 2019 | Weight factor (wf) | ∈ [0, 1] | $O(Nm(1+4T))$ | 250 | 30,000 |
| | | r | ∈ [-1, 1] | | | |
| | | μ | ∈ [0, 1] | | | |
| GWO | 2014 | a | linearly decrease from 2 to 0 | $O(Nm(1+T))$ | 1000 | 30,000 |
| | | Coefficient r1 and r2 | Random numbers in [0,1] | | | |
| DA | 2016 | cohesion, alignment, separation | Random numbers in [0.1996, -0.2] | $O(Nm(1+T))$ | 1000 | 30,000 |
| | | Food factor | Random numbers in [2, 0] | | | |
| | | Enemy factor | Random numbers in [0.0998, -0.1] | | | |
| | | Inertia factor (w) | Decreased from 0.9 to 0.2 | | | |
| AZOA | 2023 | Probability of crossover (PC) | 0.1 | $O(Nm(1+2T))$ | 500 | 30,000 |
| | | Stallion probability (SP) | 0.5 | | | |
| | | Crossover | mean | | | |
| HO | 2024 | T | Decreased from 0.9970 to 0.3679 | $O(Nm(1+5/2\,T))$ | 400 | 30,000 |
| ROA | 2021 | C | 0.1 | $O(Nm(1+3T))$ | 334 | 30,000 |
| WO | 2024 | P | 0.4 | $O(Nm(1+T))$ | 1000 | 30,000 |
| NRBO | 2024 | Deciding Factor (DF) | 0.6 | $O(Nm(1+T))$ | 1000 | 30,000 |
| FOX | 2022 | a | Decreased from 2 to 0 | $O(Nm(1+T))$ | 1000 | 30,000 |
| | | MinT | Decreased nonlinearly from 1 to 0. | | | |
| | | c1, c2 | 0.18, 0.82 | | | |
| mFOX | | $a_{modified}$ | Decreased from 1 to 0 | $O(Nm(1+T))$ | 1000 | 30,000 |
| | | c1, c2 | 0.18, 0.82 | | | |
| All algorithms | Population size N=30 Number of runs = 30 | | | | | |

## 3. Experimental results and discussions

In this section, the performance of the proposed mFOX algorithm is assessed on a range of numerical optimization problems. The evaluation utilizes a comprehensive set of benchmark functions, including unimodal, multimodal, and composite functions commonly applied in optimization research. Specifically, the mFOX algorithm was tested on 23 classical benchmark functions, 10 CEC2019 functions, and 12 CEC2022 functions, selected to provide a well-rounded evaluation across diverse optimization challenges. Additionally, four real-world engineering problems were employed to assess the practical efficiency of the proposed mFOX algorithm. To demonstrate its effectiveness, the mFOX algorithm is compared to twelve established algorithms, including WOA, TSA, ChOA, FDO, GWO, DA, AZOA, HO, ROA, WO, NRBO, and FOX, to validate its results. Control parameters for each competitor algorithm were configured based on the authors' recommendations in their original papers. The specific control parameters for the compared algorithms are outlined in Table 1.

The twelve competitor algorithms were selected based on their popularity and research impact. GWO, DA, and WOA, introduced between 2014 and 2016, are well-cited and widely used. ChOA, FDO, ROA, and TSA, introduced from 2019 to 2021, are also commonly applied. Recently developed AZOA, HO, and WO have quickly gained attention and been employed in various real-world applications. Additionally, the source codes for these algorithms are available in MATLAB, ensuring that all algorithms are tested on identical benchmark functions under the same computing conditions.

To ensure a fair comparison, the proposed mFOX algorithm and the competitor algorithms were tested on all the benchmark functions and engineering problems over thirty independent runs, each with a population size of 30 ($N = 30$). In this study, the maximum iteration parameter ($T$) was determined based on the function evaluations for all algorithms, as outlined in Table 1. Consequently, each metaheuristic underwent 30,000 function evaluations to account for computational complexity.

Performance metrics recorded include the average solution, standard deviation (STD), and ranked solutions during these runs. The comparison of algorithms is based on their average performance using the TR method, where techniques are ranked by their mean values, with rank 1 assigned to the technique with the smallest average. The algorithm with the lowest overall TR score is deemed the most effective [48]. Additionally, a statistical Wilcoxon rank test [49] was conducted to compare the significance of the algorithms' performance. All experiments were performed on a Windows 10 system with a 2.80 GHz CPU, 16.00 GB RAM, and MATLAB R2017a.

### 3.1. Comparison of Results Using twenty-three classical test functions

To evaluate the performance of the mFOX in terms of search space exploration, global optimum exploitation, and local minimum avoidance, a comprehensive set of twenty-three benchmark functions is utilized. These benchmark functions are transformed versions of complex mathematical optimization problems, incorporating shifts, rotations, scaling, and combinations to increase complexity and provide a rigorous test for optimization algorithms [50]. Additionally, this collection comprises seven unimodal functions (F1-F7), six multimodal functions (F8-F13), and ten fixed-dimensional multimodal functions (F14-F23). The unimodal functions, each characterized by a singular global optimum, are instrumental in assessing the algorithm's convergence speed, thereby providing valuable insights into its exploitation capabilities. Conversely, the multimodal functions, which encompass multiple optimal solutions, are employed to measure the algorithm's exploration proficiency. Additionally, the fixed-dimensional multimodal functions, distinguished by their relatively fewer local extrema and simplified dimensionality in comparison to more complex multimodal functions, offer a balanced evaluation of both exploration and exploitation. This dual challenge allows for a thorough assessment of the algorithm's efficacy in navigating both local and global search landscapes. Detailed information regarding this benchmark set is presented in Table 2.

The proposed mFOX algorithm's performance was evaluated against other algorithms listed in Table 1 on classical benchmark functions by analyzing the average (AVG) and standard deviation (STD) of the results. The comparison of these algorithms is based on their average performance values. Rankings are established through the TR method, in which algorithms are evaluated based on their average values, assigning the highest rank to the algorithm with the lowest average. As illustrated in Table 3, the outcomes achieved by mFOX consistently outperforms competing algorithms. For the unimodal functions (F1-F7), the proposed mFOX algorithm achieved the top rank in five functions (F1, F2, F3, F4, and F6) and a second rank in one function. In the case of multimodal functions (F8-F13), mFOX attained the top rank in four out of six functions (F9, F10, F11, and F12). Additionally, for the ten fixed-dimensional multimodal functions (F14-F23), mFOX secured the top rank in eight functions (F16-F23). When comparing the proposed algorithm with the original FOX optimizer, the proposed mFOX algorithm outperformed the FOX optimizer in 15 functions and achieved the same results as the FOX optimizer in seven functions (F1, F2, F3, F4, F9, F10, and F11). However, the FOX optimizer yielded better results in only one function (F7). This indicates that the mFOX algorithm shows enhanced exploitation abilities and quicker convergence speeds when tested on unimodal benchmark functions. Additionally, it demonstrates exceptional exploration abilities in multimodal test functions, outperforming other methods in avoiding local optima, efficiently exploring the search space, and balancing exploration and exploitation, thereby achieving quicker convergence to optimal solutions.

To assess the performance of the proposed mFOX algorithm, the Wilcoxon rank sum test is applied to determine significant differences between mFOX and other algorithms at a significance level of $p=0.05$. A $p$-value below 0.05 strongly rejects the null hypothesis, indicating a significant difference. Table 4 presents the Wilcoxon rank sum test results on the 23 classical benchmark functions. Values less than 0.05 are bolded to indicate significant difference. *'NaN'* signifies identical outcomes between mFOX and the compared algorithm. Table 4 shows that for most of the 23 classical benchmark functions, the $p$-values are below 0.05, indicating significant differences between the optimization results of mFOX and other algorithms. The results also demonstrate that mFOX outperforms the FOX algorithm on all test functions except F13. On F1, F2, F3, F4, F9, F10, and F11, both algorithms achieved the optimal value, showing no significant difference in performance. Additionally, similar trends are observed with other comparative algorithms, further highlighting mFOX's enhanced convergence performance relative to its peers.

Figure 5 presents the convergence curves of the mFOX algorithm alongside those of its competitor algorithms across benchmark functions F1 to F23. The best score in the convergence curves corresponds to the optimal value of the objective function achieved up to each iteration, with this value being updated iteratively based on comparisons with previous iterations. A detailed examination of the convergence curves reveals that the mFOX algorithm demonstrates superior performance in terms of convergence speed and robustness, particularly when addressing unimodal problems (F1 to F7), high-dimensional multimodal problems (F8 to F13), and fixed-dimensional multimodal functions (F14 to F23). Compared to the twelve competitor algorithms, mFOX consistently exhibits faster convergence and stronger optimization capability across these diverse problem types.

**Table 2:** Details of 23 basic benchmark functions.

| Function | Dim | Range | $f_{min}$ |
|---|---|---|---|
| **Unimodal benchmark function** | | | |
| $f1(x) = \sum_{i=1}^{Dim} x_i^2$ | 30 | $[-100,100]^{Dim}$ | 0 |
| $f2(x) = \sum_{i=1}^{Dim}|x_i| + \prod_{i=1}^{Dim}|x_i|$ | 30 | $[-10,10]^{Dim}$ | 0 |
| $f3(x) = \sum_{i=1}^{Dim}\left(\sum_{j=1}^{i} x_j\right)^2$ | 30 | $[-100,100]^{Dim}$ | 0 |
| $f4(x) = max_i\{|x_i|, 1 \leq i \leq Dim\}$ | 30 | $[-100,100]^{Dim}$ | 0 |
| $f5(x) = \sum_{i=1}^{Dim-1}[100(x_{i+1} - x_i^2)^2 + (x_i - 1)^2]$ | 30 | $[-30,30]^{Dim}$ | 0 |
| $f6(x) = \sum_{i=1}^{Dim}(x_i + 0.5)^2$ | 30 | $[-100,100]^{Dim}$ | 0 |
| $f7(x) = \sum_{i=1}^{Dim} ix_i^4 + random[0,1)$ | 30 | $[-1.28,1.28]^{Dim}$ | 0 |
| **Multi-modal benchmark function** | | | |
| $f8(x) = \sum_{i=1}^{Dim} -x_i sin\left(\sqrt{|x_i|}\right)$ | 30 | $[-500,500]^{Dim}$ | $-12569$ |
| $f9(x) = \sum_{i=1}^{Dim}[x_i^2 - 10cos(2\pi x_i) + 10]$ | 30 | $[-5.12,5.12]^{Dim}$ | 0 |
| $f10(x) = -20exp\left(-0.2\sqrt{\frac{\sum_{i=1}^{Dim} x_i^2}{Dim}}\right) - exp\left(\frac{\sum_{i=1}^{Dim} cos(2\pi x_i)}{Dim}\right) + 20 + e$ | 30 | $[-32,32]^{Dim}$ | 0 |
| $f11(x) = \frac{1}{4000}\sum_{i=1}^{Dim} x_i^2 - \prod_{i=1}^{Dim} cos\left(\frac{x_i}{\sqrt{i}}\right) + 1$ | 30 | $[-600,600]^{Dim}$ | 0 |

| Function | Dim | Range | $f_{min}$ |
|---|---|---|---|
| $f12(x) = \{10sin(\pi y_1) + \sum_{i=1}^{Dim-1}(y_i - 1)^2[1 + 10sin^2(\pi y_{i+1})] - (y_{Dim} - 1)^2\} + \sum_{i=1}^{Dim} u(x_i, 5, 100, 4)$ $y_i = 1 + (x_i + 1)/4$  $u(x_i, a, k, m) = \begin{cases} k(x_i - a)^m, x_i > a \\ 0, -a < x_i < a \\ k(-x_i - a)^m, x_i < -a \end{cases}$ | 30 | $[-50, 50]^{Dim}$ | 0 |
| $f13(x) = 0.1\{sin^2(3\pi x_i) + \sum_{i=1}^{Dim}(x_i - 1)^2[1 + sin^2(3\pi x_i + 1)] + (x_{Dim} - 1)^2[1 + sin^2(2\pi x_{Dim})]\} + \sum_{i=1}^{Dim} u(x_i, 5, 100, 4)$ | 30 | $[-50, 50]^{Dim}$ | 0 |
| **Fixed-dimension multi-modal benchmark function** | | | |
| $f14(x) = \left[\frac{1}{500} + \sum_{j=1}^{25} \frac{1}{j + \sum_{i=1}^{2}(x_i - a_{ij})^6}\right]^{-1}$ | 2 | $[-65, 65]^{Dim}$ | 1 |
| $f15(x) = \sum_{i=1}^{11}[a_i - [x_1(b_i^2 + b_i x_2)]/(b_i^2 + b_i x_3 + x_4)]^2$ | 4 | $[-5, 5]^{Dim}$ | 0.00030 |
| $f16(x) = 4x_1^2 - 2.1x_1^4 + 1/3\, x_1^6 + x_1 x_2 - 4x_2^2 + 4x_2^4$ | 2 | $[-5, 5]^{Dim}$ | -1.0316 |
| $f17(x) = (x_2 - 5.1x_1^2/4\pi^2 + 5x_1/\pi - 6)^2 + 10(1 - 1/8\pi)cos x_1 + 10$ | 2 | $[-5, 10] \times [0, 15]$ | 0.398 |
| $f18(x) = [1 + (x_1 + x_2 + 1)^2(19 - 14x_1 + 3x_1^2 - 14x_2 + 6x_1 x_2 + 3x_2^2)] \times [30 + (2x_1 - 3x_2)^2(18 - 32x_1 + 12x_1^2 + 48x_2 - 36x_1 x_2 + 27x_2^2)]$ | 2 | $[-2, 2]^{Dim}$ | 3 |
| $f19(x) = -\sum_{i=1}^{4} c_i \exp\left(-\sum_{j=1}^{3} a_{ij}(x_i - p_{ij})^2\right)$ | 3 | $[1, 3]^{Dim}$ | -3.86 |
| $f20(x) = -\sum_{i=1}^{4} c_i \exp\left(-\sum_{j=1}^{6} a_{ij}(x_i - p_{ij})^2\right)$ | 6 | $[0, 1]^{Dim}$ | -3.32 |
| $f21(x) = -\sum_{i=1}^{5}[(X - a_i)(X - a_i)^T + c_i]^{-1}$ | 4 | $[0, 10]^{Dim}$ | -10.1532 |
| $f22(x) = -\sum_{i=1}^{7}[(X - a_i)(X - a_i)^T + c_i]^{-1}$ | 4 | $[0, 10]^{Dim}$ | -10.4028 |
| $f23(x) = -\sum_{i=1}^{10}[(X - a_i)(X - a_i)^T + c_i]^{-1}$ | 4 | $[0, 10]^{Dim}$ | -10.5363 |

**Table 3:** Numerical results of the different algorithms on solving the 23 standard test function set.

| Funs | Mea. | mFOX | FOX | WOA | TSA | ChOA | FDO | NRBO | GWO | DA | WO | AZOA | HO | ROA |
|---|---|---|---|---|---|---|---|---|---|---|---|---|---|---|
| F1 | AVG | **0.00E+00** | **0.00E+00** | 1.16E-98 | 2.68E-47 | 9.30E-15 | 1.92E-05 | **0.00E+00** | 2.54E-59 | 1.06E+03 | 4.92E-285 | 1.21E-106 | 2.03E-285 | 1.45E-10 |
|  | STD | 0.00E+00 | 0.00E+00 | 6.20E-98 | 6.68E-47 | 1.93E-14 | 2.97E-05 | 0.00E+00 | 3.48E-59 | 6.82E+02 | 0.00E+00 | 6.17E-106 | 0.00E+00 | 6.18E-10 |
|  | RANK | 1 | 1 | 7 | 9 | 10 | 12 | 1 | 8 | 13 | 5 | 6 | 4 | 11 |
| F2 | AVG | **0.00E+00** | **0.00E+00** | 2.15E-68 | 7.86E-29 | 8.62E-11 | 2.78E-03 | 4.91E-285 | 9.30E-35 | 1.29E+01 | 8.73E-148 | 1.52E-56 | 2.46E-144 | 7.69E-07 |
|  | STD | 0.00E+00 | 0.00E+00 | 7.02E-68 | 1.73E-28 | 2.71E-10 | 1.44E-03 | 0.00E+00 | 1.01E-34 | 5.63E+00 | 3.17E-147 | 6.99E-56 | 1.14E-143 | 2.32E-06 |
|  | RANK | 1 | 1 | 6 | 9 | 10 | 12 | 3 | 8 | 13 | 4 | 7 | 5 | 11 |
| F3 | AVG | **0.00E+00** | **0.00E+00** | 3.58E+04 | 1.49E-11 | 9.97E-01 | 4.47E+02 | **0.00E+00** | 2.63E-14 | 1.36E+04 | 1.63E-270 | 2.60E-76 | 6.31E-292 | 1.36E-07 |
|  | STD | 0.00E+00 | 0.00E+00 | 1.33E+04 | 3.90E-11 | 4.43E+00 | 2.48E+02 | 0.00E+00 | 9.45E-14 | 7.39E+03 | 0.00E+00 | 1.40E-75 | 0.00E+00 | 7.30E-07 |
|  | RANK | 1 | 1 | 13 | 8 | 10 | 11 | 1 | 7 | 12 | 5 | 6 | 4 | 9 |
| F4 | AVG | **0.00E+00** | **0.00E+00** | 4.62E+01 | 4.83E-03 | 2.81E-03 | 9.55E+00 | 8.36E-282 | 1.51E-14 | 2.53E+01 | 3.20E-137 | 5.99E-46 | 9.11E-145 | 2.94E-07 |
|  | STD | 0.00E+00 | 0.00E+00 | 2.49E+01 | 8.45E-03 | 6.31E-03 | 2.95E+00 | 0.00E+00 | 3.20E-14 | 6.94E+00 | 1.72E-136 | 2.23E-45 | 4.84E-144 | 7.90E-07 |
|  | RANK | 1 | 1 | 13 | 10 | 9 | 11 | 3 | 7 | 12 | 5 | 6 | 4 | 8 |
| F5 | AVG | 2.79E+01 | 2.88E+01 | 2.78E+01 | 2.84E+01 | 2.88E+01 | 9.27E+01 | 2.79E+01 | 2.71E+01 | 1.44E+05 | **2.18E-02** | 2.78E+01 | 1.46E-01 | 3.92E+00 |
|  | STD | 3.27E-01 | 3.20E-02 | 4.87E-01 | 8.30E-01 | 4.27E-01 | 8.36E+01 | 7.45E-01 | 8.80E-01 | 1.08E+05 | 6.02E-02 | 5.42E-01 | 2.96E-01 | 8.60E+00 |
|  | RANK | 8 | 10 | 5 | 9 | 11 | 12 | 7 | 4 | 13 | 1 | 6 | 2 | 3 |
| F6 | AVG | **3.56E-06** | 2.85E-03 | 2.38E-01 | 3.67E+00 | 3.13E+00 | 1.29E-04 | 2.89E+00 | 6.73E-01 | 1.03E+03 | 2.33E-04 | 3.18E-01 | 1.60E-02 | 1.66E-02 |
|  | STD | 1.43E-06 | 1.07E-03 | 1.69E-01 | 5.33E-01 | 4.65E-01 | 1.25E-04 | 4.63E-01 | 3.32E-01 | 6.01E+02 | 2.57E-04 | 2.19E-01 | 1.78E-02 | 2.77E-02 |
|  | RANK | 1 | 4 | 7 | 12 | 11 | 2 | 10 | 9 | 13 | 3 | 8 | 5 | 6 |
| F7 | AVG | 1.38E-04 | **6.05E-05** | 2.00E-03 | 5.37E-03 | 6.49E-04 | 7.51E-01 | 1.71E-04 | 8.36E-04 | 3.32E-01 | 1.78E-04 | 8.55E-04 | 2.00E-04 | 2.42E-04 |
|  | STD | 1.22E-04 | 4.18E-05 | 2.03E-03 | 2.47E-03 | 5.39E-04 | 2.73E-01 | 1.28E-04 | 4.02E-04 | 1.78E-01 | 1.66E-04 | 5.10E-04 | 1.62E-04 | 3.45E-04 |
|  | RANK | 2 | 1 | 10 | 11 | 7 | 13 | 3 | 8 | 12 | 4 | 9 | 5 | 6 |
| F8 | AVG | -7.57E+03 | -6.99E+03 | -1.12E+04 | -6.22E+03 | -5.75E+03 | -6.43E+03 | -4.96E+03 | -6.11E+03 | -5.60E+03 | -1.26E+04 | -7.12E+03 | -1.26E+04 | **-1.26E+04** |
|  | STD | 7.84E+02 | 5.98E+02 | 1.49E+03 | 4.87E+02 | 6.57E+01 | 6.41E+02 | 6.90E+02 | 4.35E+02 | 6.71E+02 | 2.16E-01 | 9.64E+02 | 5.57E-02 | 2.71E-02 |
|  | RANK | 5 | 7 | 4 | 9 | 11 | 8 | 13 | 10 | 12 | 3 | 6 | 2 | 1 |
| F9 | AVG | **0.00E+00** | **0.00E+00** | 1.89E-15 | 1.59E+02 | 3.88E+00 | 2.72E+01 | **0.00E+00** | 1.10E-01 | 1.61E+02 | **0.00E+00** | **0.00E+00** | **0.00E+00** | 6.37E-13 |
|  | STD | 0.00E+00 | 0.00E+00 | 1.02E-14 | 4.26E+01 | 6.40E+00 | 1.19E+01 | 0.00E+00 | 5.90E-01 | 3.94E+01 | 0.00E+00 | 0.00E+00 | 0.00E+00 | 2.47E-12 |
|  | RANK | 1 | 1 | 7 | 12 | 10 | 11 | 1 | 9 | 13 | 1 | 1 | 1 | 8 |
| F10 | AVG | **8.88E-16** | **8.88E-16** | 4.44E-15 | 1.11E+00 | 2.00E+01 | 1.92E-01 | **8.88E-16** | 1.62E-14 | 8.78E+00 | **8.88E-16** | 1.01E-15 | **8.88E-16** | 8.01E-08 |

| Funs | Mea. | mFOX | FOX | WOA | TSA | ChOA | FDO | NRBO | GWO | DA | WO | AZOA | HO | ROA |
|---|---|---|---|---|---|---|---|---|---|---|---|---|---|---|
| | STD | 0.00E+00 | 0.00E+00 | 2.75E-15 | 1.48E+00 | 1.50E-03 | 5.04E-01 | 0.00E+00 | 3.80E-15 | 1.39E+00 | 0.00E+00 | 6.38E-16 | 0.00E+00 | 2.21E-07 |
| | RANK | 1 | 1 | 7 | 11 | 13 | 10 | 1 | 8 | 12 | 1 | 6 | 1 | 9 |
| F11 | AVG | **0.00E+00** | **0.00E+00** | 5.16E-03 | 5.80E-03 | 8.68E-03 | 2.35E-02 | **0.00E+00** | 5.05E-03 | 1.10E+01 | **0.00E+00** | **0.00E+00** | **0.00E+00** | 1.16E-10 |
| | STD | 0.00E+00 | 0.00E+00 | 2.78E-02 | 6.78E-03 | 2.32E-02 | 2.52E-02 | 0.00E+00 | 1.39E-02 | 4.21E+00 | 0.00E+00 | 0.00E+00 | 0.00E+00 | 3.92E-10 |
| | RANK | 1 | 1 | 9 | 10 | 11 | 12 | 1 | 8 | 13 | 1 | 1 | 1 | 7 |
| F12 | AVG | **4.72E-07** | 6.23E-05 | 1.35E-02 | 6.63E+00 | 4.24E-01 | 8.34E-01 | 2.36E-01 | 3.68E-02 | 6.27E+02 | 1.79E-06 | 1.53E-02 | 3.25E-04 | 2.79E-04 |
| | STD | 2.82E-07 | 1.82E-05 | 7.67E-03 | 3.35E+00 | 2.22E-01 | 9.73E-01 | 6.02E-02 | 1.57E-02 | 3.05E+03 | 2.45E-06 | 2.90E-02 | 6.24E-04 | 5.99E-04 |
| | RANK | 1 | 3 | 6 | 12 | 10 | 11 | 9 | 8 | 13 | 2 | 7 | 5 | 4 |
| F13 | AVG | 4.43E-02 | 4.02E-01 | 4.11E-01 | 2.77E+00 | 2.85E+00 | 2.46E-02 | 2.10E+00 | 6.07E-01 | 4.46E+04 | **8.86E-06** | 1.27E+00 | 3.91E-03 | 1.08E-02 |
| | STD | 2.16E-01 | 1.01E+00 | 2.43E-01 | 5.11E-01 | 1.08E-01 | 4.72E-02 | 4.19E-01 | 2.28E-01 | 1.02E+05 | 1.42E-05 | 5.45E-01 | 8.83E-03 | 2.35E-02 |
| | RANK | 5 | 6 | 7 | 11 | 12 | 4 | 10 | 8 | 13 | 1 | 9 | 2 | 3 |
| F14 | AVG | 1.40E+00 | 1.21E+01 | 2.76E+00 | 8.76E+00 | 9.98E-01 | 1.53E+00 | 2.18E+00 | 4.06E+00 | 1.03E+00 | **9.98E-01** | 1.46E+00 | 9.98E-01 | 1.52E+00 |
| | STD | 9.39E-01 | 2.18E+00 | 3.23E+00 | 4.81E+00 | 7.69E-05 | 7.57E-01 | 2.45E+00 | 4.26E+00 | 1.78E-01 | 3.79E-16 | 1.10E+00 | 3.79E-13 | 1.82E+00 |
| | RANK | 5 | 13 | 10 | 12 | 3 | 8 | 9 | 11 | 4 | 1 | 6 | 2 | 7 |
| F15 | AVG | 3.08E-04 | 4.27E-04 | 7.49E-04 | 6.54E-03 | 1.27E-03 | 3.19E-04 | 2.44E-03 | 2.32E-03 | 1.32E-03 | 3.33E-04 | 1.15E-03 | **3.08E-04** | 6.09E-04 |
| | STD | 6.34E-07 | 3.37E-04 | 5.05E-04 | 9.10E-03 | 2.36E-05 | 3.35E-05 | 5.98E-03 | 6.02E-03 | 4.91E-04 | 5.12E-05 | 3.59E-03 | 1.45E-07 | 4.55E-04 |
| | RANK | 2 | 5 | 7 | 13 | 9 | 3 | 12 | 11 | 10 | 4 | 8 | 1 | 6 |
| F16 | AVG | **-1.03E+00** | -1.00E+00 | -1.03E+00 | -1.03E+00 | -1.03E+00 | -1.03E+00 | **-1.03E+00** | -1.03E+00 | -1.03E+00 | -1.03E+00 | -1.03E+00 | -1.03E+00 | -1.03E+00 |
| | STD | 5.73E-16 | 1.47E-01 | 7.57E-10 | 9.49E-03 | 5.62E-06 | 1.63E-09 | 5.51E-16 | 7.01E-09 | 2.19E-05 | 4.17E-05 | 7.58E-09 | 1.17E-10 | 2.31E-04 |
| | RANK | 1 | 13 | 4 | 12 | 9 | 5 | 1 | 7 | 8 | 10 | 6 | 3 | 11 |
| F17 | AVG | **3.98E-01** | 3.98E-01 | 3.98E-01 | 3.98E-01 | 5.53E-01 | NA | **3.98E-01** | 3.98E-01 | 3.98E-01 | 4.00E-01 | **3.98E-01** | 3.98E-01 | 4.00E-01 |
| | STD | 0.00E+00 | 6.12E-11 | 2.64E-06 | 2.89E-05 | 8.33E-01 | NA | 0.00E+00 | 4.63E-05 | 1.39E-06 | 1.04E-02 | 0.00E+00 | 3.13E-09 | 4.60E-03 |
| | RANK | 1 | 4 | 7 | 9 | 12 | NA | 1 | 8 | 6 | 10 | 1 | 5 | 11 |
| F18 | AVG | **3.00E+00** | 2.10E+01 | 3.00E+00 | 2.78E+01 | 3.00E+00 | 3.00E+00 | 3.00E+00 | 3.00E+00 | 3.00E+00 | 3.00E+00 | 3.00E+00 | 3.00E+00 | 5.71E+00 |
| | STD | 1.24E-15 | 2.98E+01 | 1.13E-04 | 3.60E+01 | 4.40E-05 | 5.47E-10 | 2.31E-15 | 1.18E-05 | 1.24E-04 | 3.72E-14 | 5.08E-15 | 7.59E-10 | 8.11E+00 |
| | RANK | 1 | 12 | 10 | 13 | 9 | 5 | 2 | 7 | 8 | 4 | 3 | 6 | 11 |
| F19 | AVG | **-3.86E+00** | -3.86E+00 | -3.86E+00 | -3.86E+00 | -3.85E+00 | -3.86E+00 | **-3.86E+00** | -3.86E+00 | -3.86E+00 | -3.86E+00 | -3.86E+00 | -3.86E+00 | -3.76E+00 |
| | STD | 2.40E-15 | 1.56E-07 | 5.31E-03 | 5.80E-05 | 1.19E-03 | 4.45E-04 | 2.53E-15 | 1.61E-03 | 6.48E-05 | 1.27E-12 | 2.53E-15 | 7.26E-08 | 8.50E-02 |
| | RANK | 1 | 6 | 11 | 8 | 12 | 9 | 1 | 10 | 7 | 4 | 3 | 5 | 13 |
| F20 | AVG | **-3.32E+00** | -3.25E+00 | -3.22E+00 | -3.27E+00 | -2.40E+00 | -3.32E+00 | -3.23E+00 | -3.27E+00 | -3.25E+00 | -3.26E+00 | -3.27E+00 | -3.26E+00 | -2.77E+00 |

| Funs | Mea. | mFOX | FOX | WOA | TSA | ChOA | FDO | NRBO | GWO | DA | WO | AZOA | HO | ROA |
|---|---|---|---|---|---|---|---|---|---|---|---|---|---|---|
| | STD | 6.31E-09 | 5.87E-02 | 1.31E-01 | 6.86E-02 | 4.32E-01 | 5.06E-05 | 6.91E-02 | 6.74E-02 | 7.38E-02 | 5.93E-02 | 5.93E-02 | 6.73E-02 | 3.48E-01 |
| | RANK | 1 | 9 | 11 | 3 | 13 | 2 | 10 | 4 | 8 | 7 | 5 | 6 | 12 |
| F21 | AVG | **-1.02E+01** | -5.23E+00 | -7.23E+00 | -6.87E+00 | -2.78E+00 | -9.73E+00 | -9.13E+00 | -9.82E+00 | -8.19E+00 | -1.02E+01 | -7.71E+00 | -1.02E+01 | -1.01E+01 |
| | STD | 1.10E-09 | 9.15E-01 | 2.93E+00 | 3.29E+00 | 2.07E+00 | 1.59E+00 | 2.20E+00 | 1.26E+00 | 2.57E+00 | 1.82E-09 | 3.06E+00 | 5.27E-06 | 1.43E-02 |
| | RANK | 1 | 12 | 10 | 11 | 13 | 6 | 7 | 5 | 8 | 2 | 9 | 3 | 4 |
| F22 | AVG | **-1.04E+01** | -5.44E+00 | -8.67E+00 | -7.45E+00 | -4.04E+00 | -1.04E+01 | -8.74E+00 | -1.04E+01 | -7.51E+00 | -1.04E+01 | -7.33E+00 | -1.04E+01 | -1.04E+01 |
| | STD | 3.07E-10 | 1.33E+00 | 2.65E+00 | 3.43E+00 | 1.79E+00 | 6.59E-03 | 2.60E+00 | 2.96E-04 | 2.94E+00 | 1.48E-09 | 3.33E+00 | 6.02E-06 | 3.09E-02 |
| | RANK | 1 | 12 | 8 | 10 | 13 | 5 | 7 | 4 | 9 | 2 | 11 | 3 | 6 |
| F23 | AVG | **-1.05E+01** | -5.67E+00 | -8.68E+00 | -7.48E+00 | -4.39E+00 | -1.02E+01 | -8.59E+00 | -1.03E+01 | -7.61E+00 | -1.05E+01 | -8.54E+00 | -1.05E+01 | -1.05E+01 |
| | STD | 3.98E-10 | 1.62E+00 | 2.69E+00 | 3.62E+00 | 1.54E+00 | 1.34E+00 | 2.66E+00 | 1.46E+00 | 3.21E+00 | 4.17E-09 | 3.32E+00 | 5.60E-06 | 3.13E-02 |
| | RANK | 1 | 12 | 7 | 11 | 13 | 6 | 8 | 5 | 10 | 2 | 9 | 3 | 4 |

**Table 4:** Wilcoxon signed-rank test *p*-values of mFOX at a 5% significance level for 23 classical benchmark functions with compared algorithms.

| Compared Algorithms | Objective Function Type | | | | | | | | | | | |
|---|---|---|---|---|---|---|---|---|---|---|---|---|
| | **F1** | **F2** | **F3** | **F4** | **F5** | **F6** | **F7** | **F8** | **F9** | **F10** | **F11** | **F12** |
| vs. FOX | *NaN* | *NaN* | *NaN* | *NaN* | **1.7E-06** | **1.7E-06** | **4.7E-03** | **6.4E-03** | *NaN* | *NaN* | *NaN* | **1.7E-06** |
| vs. WOA | **1.7E-06** | **1.7E-06** | **1.7E-06** | **1.7E-06** | 3.2E-01 | **1.7E-06** | **1.7E-06** | **2.4E-06** | 3.2E-01 | **3.4E-05** | 3.2E-01 | **1.7E-06** |
| vs. TSA | **1.7E-06** | **1.7E-06** | **1.7E-06** | **1.7E-06** | **1.2E-03** | **1.7E-06** | **1.7E-06** | **9.3E-06** | **1.7E-06** | **1.2E-06** | **9.8E-04** | **1.7E-06** |
| vs. ChOA | **1.7E-06** | **1.7E-06** | **1.7E-06** | **1.7E-06** | **2.2E-05** | **1.7E-06** | **2.2E-05** | **1.7E-06** | **1.7E-06** | **1.7E-06** | **3.8E-06** | **1.7E-06** |
| vs. FDO | **1.7E-06** | **1.7E-06** | **1.7E-06** | **1.7E-06** | **2.9E-06** | **1.7E-06** | **1.7E-06** | **3.1E-05** | **1.7E-06** | **1.7E-06** | **1.7E-06** | **1.7E-06** |
| vs. GWO | **1.7E-06** | **1.7E-06** | **1.7E-06** | **1.7E-06** | **1.1E-04** | **1.7E-06** | **1.7E-06** | **2.4E-06** | 5.9E-02 | **7.0E-07** | **4.3E-02** | **1.7E-06** |
| vs. DA | **1.7E-06** | **1.7E-06** | **1.7E-06** | **1.7E-06** | **1.7E-06** | **1.7E-06** | **1.7E-06** | **1.7E-06** | 1.7E-06 | **1.7E-06** | **1.7E-06** | **1.7E-06** |
| vs. AZOA | **1.7E-06** | **1.7E-06** | **1.7E-06** | **1.7E-06** | 5.2E-01 | **1.7E-06** | **2.6E-06** | **3.9E-02** | *NaN* | 3.2E-01 | *NaN* | **1.7E-06** |
| vs. HO | **1.2E-05** | **1.7E-06** | **2.6E-06** | **1.7E-06** | **1.7E-06** | **1.9E-06** | 2.5E-01 | **1.7E-06** | *NaN* | *NaN* | *NaN* | **3.5E-06** |
| vs. ROA | **1.7E-06** | **1.7E-06** | **1.7E-06** | **1.7E-06** | **3.2E-06** | **1.7E-06** | 5.2E-01 | **1.7E-06** | **1.8E-02** | **1.7E-06** | **8.8E-05** | **1.9E-06** |
| vs. WO | **5.1E-03** | **1.7E-06** | **2.9E-04** | **1.7E-06** | **1.7E-06** | **1.7E-06** | 4.0E-01 | **1.7E-06** | *NaN* | *NaN* | *NaN* | **2.1E-02** |
| vs. NRBO | *NaN* | 1.7E-06 | *NaN* | **1.7E-06** | 7.7E-01 | **1.7E-06** | 5.9E-01 | **1.9E-06** | *NaN* | *NaN* | *NaN* | **1.7E-06** |
| **Compared Algorithms** | **Objective Function Type** | | | | | | | | | | | |
| | **F13** | **F14** | **F15** | **F16** | **F17** | **F18** | **F19** | **F20** | **F21** | **F22** | **F23** | |
| vs. FOX | 1.0E-01 | **1.7E-06** | **2.1E-03** | **1.7E-06** | **1.7E-06** | **1.7E-06** | **1.7E-06** | **1.7E-06** | **1.7E-06** | **1.7E-06** | **1.7E-06** | |
| vs. WOA | **2.2E-05** | **5.3E-03** | **1.7E-06** | **1.7E-06** | **1.7E-06** | **1.7E-06** | **1.7E-06** | **1.7E-06** | **1.7E-06** | **1.7E-06** | **1.7E-06** | |
| vs. TSA | **1.7E-06** | **1.7E-06** | **5.2E-06** | **1.7E-06** | **1.7E-06** | **1.7E-06** | **1.7E-06** | **1.7E-06** | **1.7E-06** | **1.7E-06** | **1.7E-06** | |
| vs. ChOA | **1.7E-06** | 9.8E-01 | **1.7E-06** | **1.7E-06** | **1.7E-06** | **1.7E-06** | **1.7E-06** | **1.7E-06** | **1.7E-06** | **1.7E-06** | **1.7E-06** | |
| vs. FDO | **4.3E-02** | **2.7E-02** | **2.8E-03** | **7.2E-06** | *NaN* | **1.7E-06** | **1.7E-06** | **1.7E-06** | **1.7E-06** | **1.7E-06** | **1.7E-06** | |
| vs. GWO | **7.0E-06** | **4.7E-03** | 4.3E-01 | **1.7E-06** | **1.7E-06** | **1.7E-06** | **1.7E-06** | **1.7E-06** | **1.7E-06** | **1.7E-06** | **1.7E-06** | |
| vs. DA | **1.7E-06** | 7.0E-01 | **1.7E-06** | **4.7E-06** | **1.8E-05** | **1.6E-05** | **1.7E-06** | **1.9E-06** | **1.7E-06** | **1.7E-06** | **1.7E-06** | |
| vs. AZOA | **1.7E-06** | 5.0E-01 | 9.3E-01 | **1.7E-06** | *NaN* | **2.5E-05** | **3.0E-05** | **4.7E-02** | **3.4E-02** | **2.2E-02** | 7.5E-01 | |
| vs. HO | 6.3E-01 | 8.7E-01 | 3.0E-01 | **2.6E-06** | **1.7E-06** | **1.7E-06** | **1.7E-06** | **1.7E-06** | **1.7E-06** | **1.7E-06** | **1.7E-06** | |
| vs. ROA | 6.9E-01 | 1.9E-01 | **1.7E-06** | **1.7E-06** | **1.7E-06** | **1.7E-06** | **1.7E-06** | **1.7E-06** | **1.7E-06** | **1.7E-06** | **1.7E-06** | |
| vs. WO | **3.2E-02** | 6.3E-01 | **1.1E-05** | **1.1E-04** | **7.4E-03** | **1.7E-06** | **2.7E-06** | **7.7E-03** | 2.6E-01 | 1.7E-01 | 8.1E-01 | |
| vs. NRBO | **1.7E-06** | 1.2E-01 | **2.8E-03** | 4.4E-01 | *NaN* | **1.5E-02** | **3.9E-02** | **1.7E-06** | **1.9E-06** | **5.2E-06** | **1.7E-06** | |

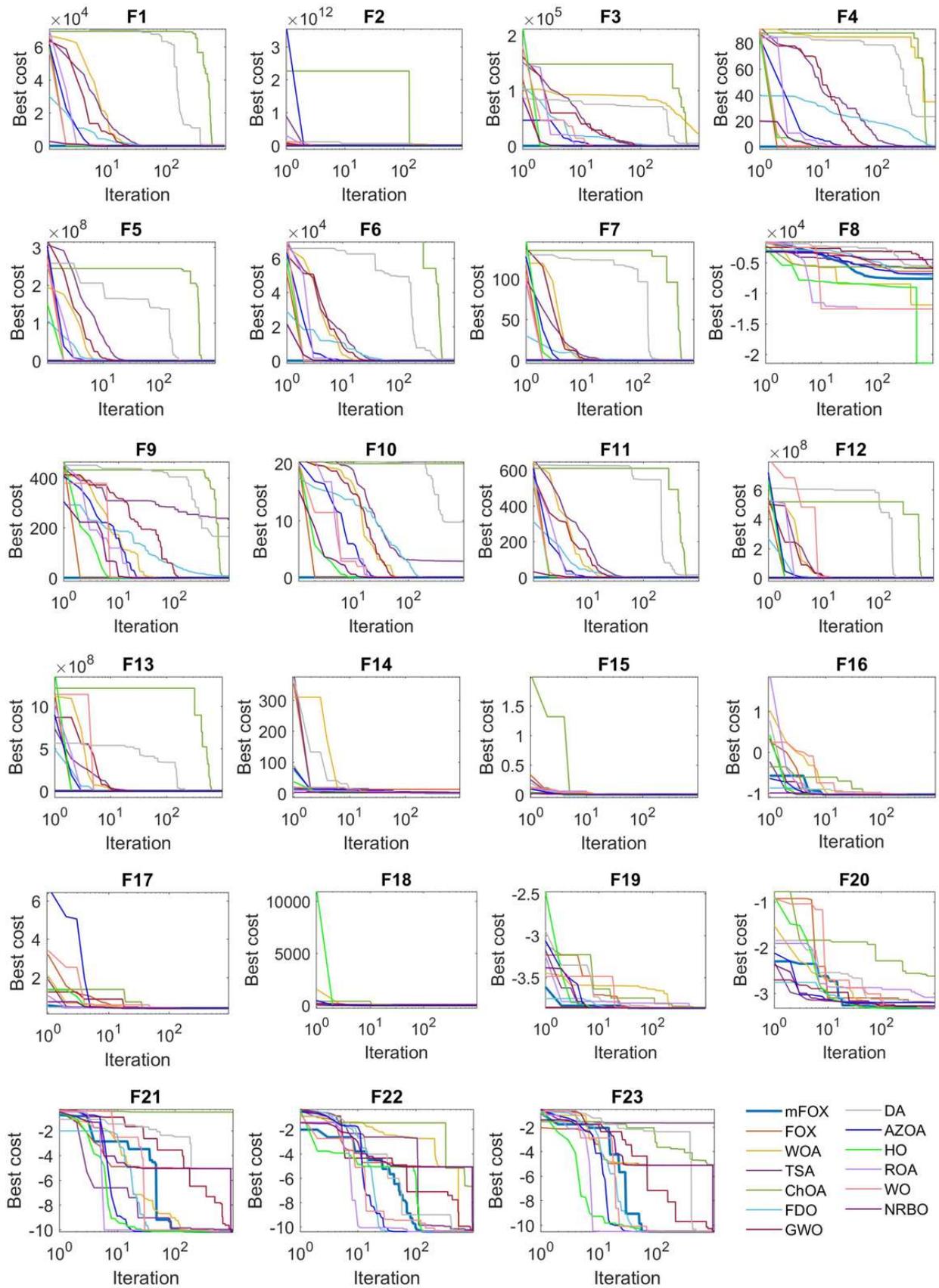

**Figure 5.** Convergence behavior analysis for the 23 classical benchmark test problems.

## 3.2. Comparison of Results Using CEC2019 Test Functions

This experiment assesses the effectiveness of the mFOX algorithm in solving a difficult set of functions known as CEC2019. Table 3 outlines the ten functions of CEC2019, including their respective dimensions and search ranges. The CEC2019 benchmark comprises ten single-objective optimization functions, which present significantly higher complexity compared to the classical functions examined in this study. These problems are notably more challenging, characterized by multilocal optima, making them ideal for evaluating an algorithm's accuracy, reliability, convergence speed, and ability to escape local optima. Further details on these functions are available in the research by [51]. The same algorithms, with parameters specified in Table 1, were applied to this benchmark. Table 6 compares the performance of mFOX with other competing algorithms. Both the mFOX algorithm and the competing methods were executed with a population size of 30, repeated across 30 independent runs. The number of iterations for each algorithm was set according to the function evaluations, as specified in Table 1.

Despite the challenges posed by these test functions, which involve randomly shifting the global optimum before each run, sometimes placing it near the boundaries of the search space, and applying rotations to the functions, mFOX demonstrates superior performance compared to other methods. This suggests that the proposed mFOX is highly effective in handling complex real-world optimization problems. Table 6 demonstrates that the mFOX algorithm outperforms other methods when tested on CEC2019 benchmark functions. It consistently produces better results in terms of both accuracy and convergence speed. mFOX effectively avoids getting trapped in local minima, enabling it to explore the entire search space and identify the best possible solutions. Using a random number to divide exploration and exploitation allows mFOX to find accurate solutions without becoming confined to suboptimal regions. As a result, mFOX, with the exception of F1, outperformed the original FOX algorithm, demonstrating superior performance across the remaining nine benchmark functions. In comparison to other competing algorithms, mFOX demonstrated exceptional performance, securing top ranks in functions 3, 5, 6, 8, 9, and 10. Although it did not claim the first position for functions 1, 2, 4, and 7, it still performed remarkably well, achieving 2nd place for functions 1, 2 and 4, and obtained 3rd place for function 7 among the 12 algorithms evaluated.

To assess the differences in results generated by the mFOX algorithm compared to competing algorithms across the ten functions of CEC2019, Table 7 displays the outcomes of the Wilcoxon rank-sum test, which compares mFOX with twelve other algorithms. The findings reveal that for the majority of the ten CEC2019 benchmark functions, the *p*-values are below 0.05, indicating significant differences in the optimization results between mFOX and the other algorithms. Additionally, the results show that mFOX outperforms the FOX algorithm on all test functions, except for the first one.

To facilitate a comprehensive evaluation of the effectiveness of the mFOX algorithm in addressing the CEC2019 benchmark functions, Figure 6 illustrates the convergence curves. The analysis of these curves indicates that mFOX demonstrates a competitive convergence rate relative to other optimization methods across a wide range of scenarios. This suggests that mFOX is capable of effectively converging toward optimal solutions in most cases, highlighting its potential as a robust algorithm for tackling complex optimization challenges.

**Table 5:** Description of CEC2019 benchmark.

| No | Functions | Dim | [lb, ub] | $F_{min}$ |
|---|---|---|---|---|
| 1 | Storn's Chebyshev Polynomial Fitting Problem | 9 | [− 8192, 8192] | 1 |
| 2 | Inverse Hilbert Matrix Problem | 16 | [− 16384, 16384] | 1 |
| 3 | Lennard-Jones Minimum Energy Cluster | 18 | [− 4,4] | 1 |
| 4 | Rastrigin's Function | 10 | [− 100,100] | 1 |
| 5 | Griewangk's Function | 10 | [− 100,100] | 1 |

| 6 | Weierstrass Function | 10 | [− 100,100] | 1 |
| 7 | Modified Schwefel's Function | 10 | [− 100,100] | 1 |
| 8 | Expanded Schaffer's F6 Function | 10 | [− 100,100] | 1 |
| 9 | Happy Cat Function | 10 | [− 100,100] | 1 |
| 10 | Ackley Function | 10 | [− 100,100] | 1 |

Table 6: Numerical results of mFOX against different algorithms on solving the CEC2019 test function set.

| Funs | Mea. | mFOX | FOX | WOA | TSA | ChOA | FDO | GWO | DA | AZOA | HO | ROA | WO | NRBO |
|---|---|---|---|---|---|---|---|---|---|---|---|---|---|---|
| cec1 | AVG | 4.418E+04 | **4.328E+04** | 2.091E+10 | 1.428E+08 | 1.855E+09 | 2.565E+08 | 1.030E+08 | 4.579E+10 | 1.900E+05 | 5.076E+04 | 1.614E+05 | 4.701E+04 | 7.145E+04 |
| | STD | 2.923E+03 | 1.627E+03 | 2.695E+10 | 3.765E+08 | 4.347E+09 | 2.159E+08 | 2.507E+08 | 4.297E+10 | 3.186E+05 | 6.832E+03 | 1.565E+05 | 3.379E+03 | 1.094E+05 |
| | RANK | 2 | 1 | 12 | 9 | 11 | 10 | 8 | 13 | 7 | 4 | 6 | 3 | 5 |
| cec2 | AVG | 1.734E+01 | 1.734E+01 | 1.736E+01 | 1.838E+01 | 1.739E+01 | **1.734E+01** | 1.737E+01 | 8.176E+01 | 1.734E+01 | 1.740E+01 | 1.779E+01 | 1.734E+01 | 1.741E+01 |
| | STD | 1.925E-07 | 1.696E-04 | 5.632E-02 | 6.055E-01 | 1.394E-02 | 1.014E-09 | 8.123E-02 | 9.513E+01 | 1.928E-05 | 4.795E-02 | 3.190E-01 | 5.455E-05 | 1.036E-01 |
| | RANK | 2 | 5 | 6 | 12 | 8 | 1 | 7 | 13 | 3 | 9 | 11 | 4 | 10 |
| cec3 | AVG | **1.270E+01** | 1.270E+01 | 1.270E+01 | 1.270E+01 | 1.270E+01 | 1.270E+01 | 1.270E+01 | 1.270E+01 | 1.270E+01 | 1.270E+01 | 1.270E+01 | 1.270E+01 | 1.270E+01 |
| | STD | 9.457E-13 | 4.945E-09 | 6.934E-07 | 8.184E-04 | 5.010E-06 | 1.254E-11 | 6.355E-07 | 4.473E-04 | 2.798E-06 | 8.459E-11 | 1.108E-03 | 4.568E-07 | 1.452E-05 |
| | RANK | 1 | 4 | 7 | 12 | 10 | 2 | 6 | 11 | 8 | 3 | 13 | 5 | 9 |
| cec4 | AVG | 4.378E+01 | 1.040E+03 | 2.884E+02 | 4.837E+03 | 4.505E+03 | **3.576E+01** | 4.923E+01 | 3.527E+02 | 3.311E+02 | 2.407E+02 | 6.204E+03 | 5.513E+01 | 1.281E+03 |
| | STD | 3.133E+01 | 5.168E+02 | 8.565E+01 | 4.184E+03 | 2.893E+03 | 1.527E+01 | 2.343E+01 | 3.641E+02 | 2.708E+02 | 1.141E+02 | 4.065E+03 | 3.113E+01 | 1.166E+03 |
| | RANK | 2 | 9 | 6 | 12 | 11 | 1 | 3 | 8 | 7 | 5 | 13 | 4 | 10 |
| cec5 | AVG | **1.121E+00** | 5.412E+00 | 1.894E+00 | 2.915E+00 | 2.646E+00 | 1.173E+00 | 1.381E+00 | 1.709E+00 | 1.524E+00 | 1.410E+00 | 2.950E+00 | 1.347E+00 | 2.029E+00 |
| | STD | 6.310E-02 | 1.276E+00 | 3.956E-01 | 1.067E+00 | 4.959E-01 | 8.864E-02 | 2.496E-01 | 2.629E-01 | 4.773E-01 | 1.998E-01 | 5.501E-01 | 3.303E-01 | 2.118E-01 |
| | RANK | 1 | 13 | 8 | 11 | 10 | 2 | 4 | 7 | 6 | 5 | 12 | 3 | 9 |
| cec6 | AVG | **2.777E+00** | 3.407E+00 | 9.293E+00 | 1.077E+01 | 1.066E+01 | 9.862E+00 | 1.057E+01 | 9.923E+00 | 9.102E+00 | 6.748E+00 | 1.030E+01 | 1.058E+01 | 9.262E+00 |
| | STD | 1.112E+00 | 1.151E+00 | 1.215E+00 | 5.625E-01 | 7.005E-01 | 9.197E-01 | 6.388E-01 | 1.126E+00 | 1.297E+00 | 9.402E-01 | 7.807E-01 | 1.095E+00 | 9.301E-01 |
| | RANK | 1 | 2 | 6 | 13 | 12 | 7 | 10 | 8 | 4 | 3 | 9 | 11 | 5 |
| cec7 | AVG | 1.448E+02 | 3.767E+02 | 5.077E+02 | 5.490E+02 | 9.290E+02 | 8.009E+01 | 3.277E+02 | 5.340E+02 | 2.455E+02 | 1.240E+02 | 9.563E+02 | 7.582E+02 | 4.067E+02 |
| | STD | 1.543E+02 | 2.711E+02 | 2.338E+02 | 1.655E+02 | 1.557E+02 | 9.365E+01 | 2.867E+02 | 2.033E+02 | 2.114E+02 | 1.364E+02 | 2.935E+02 | 3.568E+02 | 1.658E+02 |
| | RANK | 3 | 6 | 8 | 10 | 12 | 1 | 5 | 9 | 4 | 2 | 13 | 11 | 7 |
| cec8 | AVG | **4.053E+00** | 5.793E+00 | 5.767E+00 | 6.104E+00 | 6.746E+00 | 4.647E+00 | 5.047E+00 | 5.805E+00 | 4.845E+00 | 4.753E+00 | 6.385E+00 | 5.832E+00 | 5.079E+00 |
| | STD | 5.763E-01 | 3.942E-01 | 6.280E-01 | 6.081E-01 | 1.595E-01 | 4.650E-01 | 8.123E-01 | 5.769E-01 | 6.311E-01 | 6.219E-01 | 4.352E-01 | 7.056E-01 | 6.121E-01 |
| | RANK | 1 | 8 | 7 | 11 | 13 | 2 | 5 | 9 | 4 | 3 | 12 | 10 | 6 |
| cec9 | AVG | **2.347E+00** | 2.352E+00 | 4.837E+00 | 4.459E+02 | 3.948E+02 | 2.521E+00 | 4.423E+00 | 3.887E+00 | 3.090E+00 | 3.967E+00 | 8.659E+02 | 2.839E+00 | 2.077E+01 |
| | STD | 3.449E-03 | 1.067E-02 | 9.398E-01 | 5.887E+02 | 1.989E+02 | 8.538E-02 | 1.007E+00 | 7.286E-01 | 4.106E-01 | 7.244E-01 | 4.605E+02 | 2.655E-01 | 1.334E+01 |
| | RANK | 1 | 2 | 9 | 12 | 11 | 3 | 8 | 6 | 5 | 7 | 13 | 4 | 10 |
| cec10 | AVG | **1.734E+01** | 1.999E+01 | 2.021E+01 | 2.042E+01 | 2.045E+01 | 1.943E+01 | 2.042E+01 | 2.036E+01 | 1.925E+01 | 2.000E+01 | 2.042E+01 | 1.975E+01 | 2.006E+01 |
| | STD | 5.302E+00 | 2.783E-02 | 9.861E-02 | 8.424E-02 | 6.429E-02 | 3.611E+00 | 9.527E-02 | 1.231E-01 | 2.759E+00 | 1.357E-02 | 1.442E-01 | 3.667E+00 | 7.580E-01 |
| | RANK | 1 | 5 | 8 | 12 | 13 | 3 | 11 | 9 | 2 | 6 | 10 | 4 | 7 |

**Table 7:** Wilcoxon signed-rank test *p*-values at a 5% significance level for CEC2019 benchmark functions.

| Compared Algorithms | Objective Function Type | | | | | | | | | |
|---|---|---|---|---|---|---|---|---|---|---|
| | cec1 | cec2 | cec3 | cec4 | cec5 | cec6 | cec7 | cec8 | cec9 | cec10 |
| vs. FOX | 2.54E-01 | 1.73E-06 | 1.73E-06 | 1.73E-06 | 1.73E-06 | 3.68E-02 | 1.11E-03 | 1.73E-06 | 2.70E-02 | 3.00E-02 |
| vs. WOA | 1.73E-06 | 1.73E-06 | 1.73E-06 | 1.73E-06 | 1.73E-06 | 1.73E-06 | 6.98E-06 | 2.13E-06 | 1.73E-06 | 1.73E-06 |
| vs. TSA | 3.18E-06 | 1.73E-06 | 1.73E-06 | 1.73E-06 | 1.73E-06 | 1.73E-06 | 2.35E-06 | 1.92E-06 | 1.73E-06 | 1.73E-06 |
| vs. ChOA | 1.73E-06 | 1.73E-06 | 1.73E-06 | 1.73E-06 | 1.73E-06 | 1.73E-06 | 1.73E-06 | 1.73E-06 | 1.73E-06 | 1.73E-06 |
| vs. FDO | 1.73E-06 | 4.73E-06 | 1.20E-03 | 4.65E-01 | 4.49E-02 | 1.73E-06 | 8.97E-02 | 5.71E-04 | 1.73E-06 | 6.32E-05 |
| vs. GWO | 2.35E-06 | 1.73E-06 | 1.73E-06 | 2.89E-01 | 1.13E-05 | 1.73E-06 | 8.22E-03 | 5.31E-05 | 1.73E-06 | 1.73E-06 |
| vs. DA | 1.73E-06 | 1.73E-06 | 1.73E-06 | 8.47E-06 | 1.73E-06 | 1.73E-06 | 5.22E-06 | 1.73E-06 | 1.73E-06 | 1.73E-06 |
| vs. AZOA | 4.28E-01 | 3.11E-05 | 4.22E-02 | 1.73E-06 | 3.88E-06 | 1.73E-06 | 5.19E-02 | 2.22E-04 | 1.73E-06 | 5.67E-03 |
| vs. HO | 3.06E-04 | 1.73E-06 | 4.73E-06 | 1.92E-06 | 3.18E-06 | 1.73E-06 | 6.14E-01 | 1.48E-03 | 1.73E-06 | 2.83E-04 |
| vs. ROA | 1.73E-06 | 1.73E-06 | 1.73E-06 | 1.73E-06 | 1.73E-06 | 1.73E-06 | 1.73E-06 | 1.73E-06 | 1.73E-06 | 1.73E-06 |
| vs. WO | 2.96E-03 | 1.73E-06 | 2.41E-03 | 1.59E-01 | 2.26E-03 | 1.73E-06 | 4.29E-06 | 2.60E-06 | 1.73E-06 | 2.16E-05 |
| vs. NRBO | 7.97E-01 | 1.73E-06 | 1.73E-06 | 1.73E-06 | 1.73E-06 | 1.73E-06 | 1.97E-05 | 1.02E-05 | 1.73E-06 | 1.73E-06 |

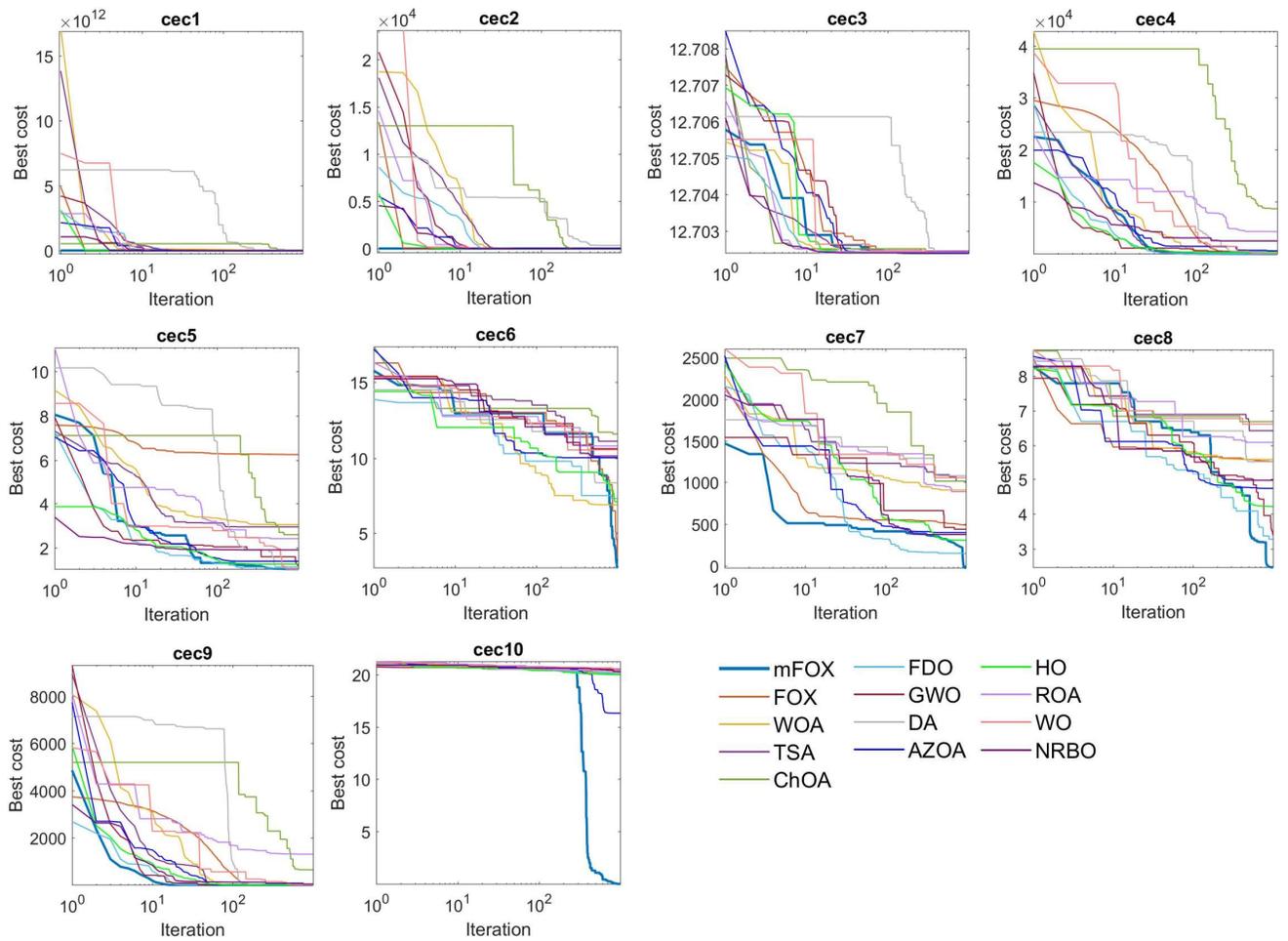

**Figure 6.** Convergence behavior analysis for the CEC2019 benchmark test problems.

Table 8 presents the running time in seconds for mFOX and the compared algorithms on each CEC2019 function, based on 30,000 function evaluations as shown in Table 1. The results, based on the total running time across all ten functions, show that the proposed mFOX algorithm outperforms FOX, ChOA, FDO, GWO, DA, HO, and NRBO in terms of speed. Among these algorithms, DA is the slowest, followed by FDO as the second slowest.

**Table 8:** Execution Time (seconds) of mFOX and Compared Algorithms for CEC2019 Functions.

| Fun. | mFOX | FOX | WOA | TSA | ChOA | FDO | GWO | DA | AZOA | HO | ROA | WO | NRBO |
|---|---|---|---|---|---|---|---|---|---|---|---|---|---|
| **cec1** | 8.80 | 8.74 | 8.28 | 7.87 | 8.98 | 138.65 | 8.26 | 28.44 | 6.49 | 11.57 | 7.73 | 7.96 | 8.32 |
| **cec2** | 0.21 | 0.37 | 0.16 | 0.12 | 1.71 | 1.38 | 0.13 | 21.62 | 0.46 | 1.79 | 0.07 | 0.15 | 0.55 |
| **cec3** | 0.28 | 0.44 | 0.22 | 0.21 | 2.29 | 2.88 | 0.21 | 22.64 | 0.50 | 1.81 | 0.13 | 0.22 | 0.64 |
| **cec4** | 0.26 | 0.41 | 0.19 | 0.15 | 1.37 | 1.91 | 0.16 | 16.09 | 0.53 | 1.69 | 0.11 | 0.17 | 0.61 |
| **cec5** | 0.27 | 0.40 | 0.20 | 0.16 | 1.36 | 2.06 | 0.16 | 19.89 | 0.54 | 1.69 | 0.11 | 0.18 | 0.62 |
| **cec6** | 2.71 | 2.82 | 2.71 | 2.57 | 3.81 | 49.93 | 4.53 | 30.75 | 2.62 | 5.15 | 3.01 | 3.03 | 3.67 |
| **cec7** | 0.34 | 0.52 | 0.26 | 0.19 | 1.78 | 3.15 | 0.25 | 30.94 | 0.56 | 2.04 | 0.14 | 0.22 | 0.63 |
| **cec8** | 0.27 | 0.42 | 0.21 | 0.17 | 1.39 | 2.77 | 0.18 | 25.26 | 0.66 | 1.91 | 0.11 | 0.18 | 0.61 |
| **cec9** | 0.25 | 0.40 | 0.19 | 0.14 | 1.57 | 1.96 | 0.14 | 17.14 | 0.49 | 2.05 | 0.10 | 0.16 | 0.57 |
| **cec10** | 0.30 | 0.45 | 0.29 | 0.25 | 1.80 | 2.70 | 0.21 | 17.92 | 0.59 | 2.18 | 0.14 | 0.20 | 0.60 |
| **SUM** | 13.69 | 14.97 | 12.72 | 11.83 | 26.06 | 207.40 | 14.22 | 230.68 | 13.43 | 31.89 | 11.67 | 12.47 | 16.83 |
| **Rank** | 6 | 8 | 4 | 2 | 10 | 12 | 7 | 13 | 5 | 11 | 1 | 3 | 9 |

### 3.3. Comparison of Results Using CEC2022 Test Functions

This section presents additional experiments using the latest CEC2022 benchmark suite [52] to emphasize the strengths of the mFOX algorithm. The CEC2022 test set comprises twelve functions, categorized into four types: unimodal, multimodal, hybrid, and composite functions (as detailed in Table 9). To evaluate the algorithm's effectiveness, its performance is compared with the algorithms listed in Table 1. The results were obtained from 30 independent trials, each using a population size of 30 and 30,000 function evaluations, ensuring robust and reliable comparison. Table 10 summarizes the performance of the mFOX algorithm compared to 12 other algorithms on the CEC2022 test suite. The mFOX algorithm achieved superior results, ranking first in 7 out of 12 test functions (F1, F2, F3, F4, F7, F10, and F11). While it did not secure the top position in the remaining four functions, it demonstrated competitive performance by ranking second in F5, F8, F9, and F12, and fourth in F6. Additionally, compared to the FOX algorithm, the proposed mFOX algorithm achieved better results across all twelve CEC2022 benchmark functions. These results highlight the robustness and effectiveness of the mFOX algorithm across a wide range of benchmark functions.

To statistically validate the mFOX results, the nonparametric Wilcoxon rank-sum test was employed. Table 11 presents the *p*-values from pairwise comparisons between mFOX and other algorithms (FOX, WOA, TSA, ChOA, FDO, GWO, DA, AZOA, HO, ROA, WO, NRBO) on the CEC2022 benchmark functions. These p-values were computed at a 0.05 significance level over 30 independent runs, assessing the statistical significance of performance differences. The findings reveal that for the majority of the 12 CEC2022 benchmark functions, the *p*-values are below 0.05, indicating significant differences in optimization performance between mFOX and the other algorithms. Moreover, the results show that mFOX consistently outperforms the FOX algorithm across all test functions.

**Table 9:** Description of CEC2022 benchmark.

| Type | No | Functions | Dim | [lb, ub] | $F_{min}$ |
|---|---|---|---|---|---|
| Unimodal functions | F1 | Shifted and full Rotated Zakharov Function | 10 | [-100,100] | 300 |
| | F2 | Shifted and full Rotated Rosenbrock's Function | 10 | [-100,100] | 400 |
| Basic functions | F3 | Shifted and full Rotated Expanded Schaffer's f6 Function | 10 | [-100,100] | 600 |
| | F4 | Shifted and full Rotated Non-continuous Rastrigin's Function | 10 | [-100,100] | 800 |
| | F5 | Shifted and Rotated Levy Function | 10 | [-100,100] | 900 |
| Hybrid functions | F6 | Hybrid function 1 (N = 3) | 10 | [-100,100] | 1800 |
| | F7 | Hybrid function 2 (N = 6) | 10 | [-100,100] | 2000 |
| | F8 | Hybrid function 3 (N = 5) | 10 | [-100,100] | 2200 |
| Composite functions | F9 | Composite function 1 (N = 5) | 10 | [-100,100] | 2300 |
| | F10 | Composite function 2 (N = 4) | 10 | [-100,100] | 2400 |
| | F11 | Composite function 3 (N = 5) | 10 | [-100,100] | 2600 |
| | F12 | Composite function 4 (N = 6) | 10 | [-100,100] | 2700 |

**Table 10:** Numerical results of mFOX against different algorithms on solving the CEC2022 test function set.

| Funs | Mea. | mFOX | FOX | WOA | TSA | ChOA | FDO | GWO | DA | AZOA | HO | ROA | WO | NRBO |
|---|---|---|---|---|---|---|---|---|---|---|---|---|---|---|
| F1 | AVG | **3.00E+02** | 3.01E+02 | 2.52E+04 | 7.00E+03 | 3.16E+03 | 3.02E+02 | 2.75E+03 | 2.39E+03 | 4.06E+02 | 1.45E+03 | 9.25E+03 | 3.09E+02 | 1.27E+03 |
|  | STD | 2.85E-07 | 4.96E+00 | 1.22E+04 | 3.20E+03 | 1.92E+03 | 4.82E+00 | 2.50E+03 | 2.07E+03 | 3.84E+02 | 6.65E+02 | 1.62E+03 | 9.22E+00 | 1.07E+03 |
|  | RANK | 1 | 2 | 13 | 11 | 10 | 3 | 9 | 8 | 5 | 7 | 12 | 4 | 6 |
| F2 | AVG | **4.03E+02** | 4.18E+02 | 4.67E+02 | 5.37E+02 | 5.65E+02 | 4.16E+02 | 4.25E+02 | 4.71E+02 | 4.22E+02 | 4.53E+02 | 8.70E+02 | 4.19E+02 | 4.45E+02 |
|  | STD | 1.28E+01 | 2.54E+01 | 4.89E+01 | 1.15E+02 | 9.96E+01 | 2.76E+01 | 2.29E+01 | 9.31E+01 | 2.86E+01 | 3.98E+01 | 2.45E+02 | 2.59E+01 | 2.72E+01 |
|  | RANK | 1 | 3 | 9 | 11 | 12 | 2 | 6 | 10 | 5 | 8 | 13 | 4 | 7 |
| F3 | AVG | **6.01E+02** | 6.51E+02 | 6.39E+02 | 6.34E+02 | 6.31E+02 | 6.09E+02 | 6.02E+02 | 6.34E+02 | 6.19E+02 | 6.28E+02 | 6.45E+02 | 6.01E+02 | 6.26E+02 |
|  | STD | 1.16E+00 | 8.54E+00 | 1.57E+01 | 1.40E+01 | 9.16E+00 | 6.73E+00 | 2.58E+00 | 2.11E+01 | 1.07E+01 | 1.20E+01 | 9.66E+00 | 1.17E+00 | 9.38E+00 |
|  | RANK | 1 | 13 | 11 | 9 | 8 | 4 | 3 | 10 | 5 | 7 | 12 | 2 | 6 |
| F4 | AVG | **8.13E+02** | 8.36E+02 | 8.39E+02 | 8.48E+02 | 8.38E+02 | 8.18E+02 | 8.16E+02 | 8.44E+02 | 8.26E+02 | 8.22E+02 | 8.53E+02 | 8.25E+02 | 8.31E+02 |
|  | STD | 3.93E+00 | 9.69E+00 | 1.47E+01 | 1.89E+01 | 6.52E+00 | 7.12E+00 | 7.99E+00 | 1.75E+01 | 9.81E+00 | 6.45E+00 | 1.02E+01 | 1.50E+01 | 7.94E+00 |
|  | RANK | 1 | 8 | 10 | 12 | 9 | 3 | 2 | 11 | 6 | 4 | 13 | 5 | 7 |
| F5 | AVG | 9.12E+02 | 1.51E+03 | 1.44E+03 | 1.50E+03 | 1.25E+03 | 9.13E+02 | 9.20E+02 | 1.34E+03 | 1.07E+03 | 1.13E+03 | 1.50E+03 | **9.10E+02** | 1.05E+03 |
|  | STD | 1.88E+01 | 1.33E+02 | 2.85E+02 | 4.05E+02 | 1.36E+02 | 2.37E+01 | 5.70E+01 | 4.86E+02 | 1.51E+02 | 1.62E+02 | 2.58E+02 | 1.81E+01 | 1.16E+02 |
|  | RANK | 2 | 13 | 10 | 12 | 8 | 3 | 4 | 9 | 6 | 7 | 11 | 1 | 5 |
| F6 | AVG | 2.77E+03 | 4.47E+03 | 4.49E+03 | 1.97E+06 | 1.06E+06 | 2.26E+03 | 6.11E+03 | 1.39E+04 | 3.63E+03 | **2.09E+03** | 6.12E+06 | 2.74E+03 | 4.35E+03 |
|  | STD | 9.50E+02 | 2.12E+03 | 2.09E+03 | 1.04E+07 | 8.60E+05 | 8.89E+02 | 2.12E+03 | 1.85E+04 | 2.11E+03 | 2.52E+02 | 1.80E+07 | 1.21E+03 | 2.13E+03 |
|  | RANK | 4 | 7 | 8 | 12 | 11 | 2 | 9 | 10 | 5 | 1 | 13 | 3 | 6 |
| F7 | AVG | **2.02E+03** | 2.18E+03 | 2.08E+03 | 2.09E+03 | 2.06E+03 | 2.03E+03 | 2.03E+03 | 2.09E+03 | 2.05E+03 | 2.05E+03 | 2.11E+03 | 2.03E+03 | 2.06E+03 |
|  | STD | 8.31E+00 | 6.43E+01 | 3.88E+01 | 4.77E+01 | 1.00E+01 | 1.71E+01 | 1.12E+01 | 3.38E+01 | 2.36E+01 | 1.73E+01 | 3.34E+01 | 1.50E+01 | 1.71E+01 |
|  | RANK | 1 | 13 | 9 | 11 | 7 | 2 | 3 | 10 | 5 | 6 | 12 | 4 | 8 |
| F8 | AVG | 2.22E+03 | 2.45E+03 | 2.24E+03 | 2.28E+03 | 2.31E+03 | **2.22E+03** | 2.23E+03 | 2.25E+03 | 2.23E+03 | 2.23E+03 | 2.25E+03 | 2.22E+03 | 2.24E+03 |
|  | STD | 6.24E+00 | 1.51E+02 | 1.21E+01 | 6.38E+01 | 5.90E+01 | 4.03E+00 | 5.25E+00 | 4.00E+01 | 6.20E+00 | 5.01E+00 | 1.94E+01 | 5.66E+00 | 3.59E+01 |
|  | RANK | 2 | 13 | 7 | 11 | 12 | 1 | 4 | 10 | 5 | 6 | 9 | 3 | 8 |
| F9 | AVG | 2.53E+03 | 2.57E+03 | 2.61E+03 | 2.66E+03 | 2.57E+03 | **2.53E+03** | 2.57E+03 | 2.61E+03 | 2.53E+03 | 2.61E+03 | 2.72E+03 | 2.54E+03 | 2.58E+03 |
|  | STD | 1.98E-03 | 4.53E+01 | 5.56E+01 | 5.54E+01 | 2.02E+01 | 7.10E+00 | 2.69E+01 | 6.26E+01 | 2.63E+01 | 6.08E+01 | 8.18E+01 | 2.77E+01 | 3.18E+01 |
|  | RANK | 2 | 7 | 10 | 12 | 6 | 1 | 5 | 11 | 3 | 9 | 13 | 4 | 8 |
| F10 | AVG | **2.53E+03** | 3.06E+03 | 2.64E+03 | 2.87E+03 | 2.95E+03 | 2.57E+03 | 2.59E+03 | 2.61E+03 | 2.57E+03 | 2.55E+03 | 2.79E+03 | 2.56E+03 | 2.61E+03 |
|  | STD | 5.04E+01 | 6.38E+02 | 3.63E+02 | 4.42E+02 | 6.30E+02 | 5.80E+01 | 8.55E+01 | 1.99E+02 | 6.30E+01 | 6.49E+01 | 4.17E+02 | 5.97E+01 | 2.13E+02 |
|  | RANK | 1 | 13 | 9 | 11 | 12 | 4 | 6 | 8 | 5 | 2 | 10 | 3 | 7 |

| Funs | Mea. | mFOX | FOX | WOA | TSA | ChOA | FDO | GWO | DA | AZOA | HO | ROA | WO | NRBO |
|---|---|---|---|---|---|---|---|---|---|---|---|---|---|---|
| F11 | AVG | **2.61E+03** | 2.75E+03 | 2.91E+03 | 3.16E+03 | 3.38E+03 | 2.71E+03 | 2.78E+03 | 2.89E+03 | 2.70E+03 | 2.76E+03 | 3.26E+03 | 2.73E+03 | 2.91E+03 |
|  | STD | 5.39E+01 | 1.70E+02 | 1.92E+02 | 4.14E+02 | 2.47E+02 | 1.44E+02 | 1.26E+02 | 2.16E+02 | 1.52E+02 | 1.66E+02 | 3.55E+02 | 1.57E+02 | 2.38E+02 |
|  | RANK | 1 | 5 | 10 | 11 | 13 | 3 | 7 | 8 | 2 | 6 | 12 | 4 | 9 |
| F12 | AVG | 2.87E+03 | 2.97E+03 | 2.90E+03 | 2.94E+03 | 2.87E+03 | 2.88E+03 | 2.87E+03 | 2.88E+03 | 2.88E+03 | 2.88E+03 | 2.94E+03 | **2.86E+03** | 2.87E+03 |
|  | STD | 2.42E+00 | 6.14E+01 | 4.44E+01 | 6.69E+01 | 9.77E+00 | 1.50E+01 | 3.38E+00 | 1.80E+01 | 1.73E+01 | 2.66E+01 | 5.65E+01 | 1.07E+00 | 1.86E+01 |
|  | RANK | 2 | 13 | 10 | 11 | 4 | 8 | 3 | 9 | 6 | 7 | 12 | 1 | 5 |

**Table 11:** Wilcoxon signed-rank test *p*-values at a 5% significance level for CEC2022 test functions.

| Compared Algorithms | Objective Function Type | | | | | | | | | | | |
|---|---|---|---|---|---|---|---|---|---|---|---|---|
| | F1 | F2 | F3 | F4 | F5 | F6 | F7 | F8 | F9 | F10 | F11 | F12 |
| vs. FOX | 1.7E-06 | 8.9E-04 | 1.7E-06 | 1.7E-06 | 1.7E-06 | 1.5E-03 | 1.7E-06 | 1.7E-06 | 1.7E-06 | 3.7E-05 | 8.5E-06 | 1.7E-06 |
| vs. WOA | 1.7E-06 | 1.7E-06 | 1.7E-06 | 2.4E-06 | 1.7E-06 | 8.2E-05 | 1.9E-06 | 1.7E-06 | 1.7E-06 | 8.2E-03 | 1.7E-06 | 3.5E-06 |
| vs. TSA | 1.7E-06 | 2.6E-06 | 1.7E-06 | 1.7E-06 | 1.7E-06 | 1.7E-06 | 1.7E-06 | 7.7E-06 | 1.7E-06 | 6.3E-05 | 2.6E-06 | 2.4E-06 |
| vs. ChOA | 1.7E-06 | 2.1E-06 | 1.7E-06 | 1.7E-06 | 1.7E-06 | 1.7E-06 | 1.7E-06 | 1.7E-06 | 1.7E-06 | 8.7E-03 | 1.7E-06 | 1.4E-04 |
| vs. FDO | 1.7E-06 | 3.7E-02 | 1.7E-06 | 6.2E-04 | 9.8E-01 | 2.3E-03 | 2.3E-03 | 3.0E-01 | 1.7E-06 | 2.8E-02 | 4.5E-02 | 4.3E-06 |
| vs. GWO | 1.7E-06 | 1.4E-05 | 3.7E-02 | 1.3E-01 | 8.3E-01 | 7.7E-06 | 4.7E-06 | 3.4E-03 | 1.7E-06 | 1.7E-02 | 1.7E-06 | 8.6E-01 |
| vs. DA | 1.7E-06 | 8.5E-06 | 1.7E-06 | 2.6E-06 | 4.3E-06 | 1.2E-05 | 1.7E-06 | 2.1E-06 | 1.7E-06 | 1.1E-04 | 3.5E-06 | 3.4E-05 |
| vs. AZOA | 1.7E-06 | 9.7E-05 | 1.7E-06 | 3.1E-05 | 2.1E-06 | 1.3E-01 | 1.7E-06 | 4.1E-03 | 4.5E-01 | 7.5E-05 | 2.2E-05 | 8.2E-05 |
| vs. HO | 1.7E-06 | 2.0E-05 | 1.7E-06 | 3.1E-05 | 1.7E-06 | 2.6E-03 | 1.7E-06 | 4.1E-05 | 1.7E-06 | 2.8E-02 | 1.2E-05 | 8.9E-04 |
| vs. ROA | 1.7E-06 | 1.7E-06 | 1.7E-06 | 1.7E-06 | 1.7E-06 | 1.7E-06 | 1.7E-06 | 1.7E-06 | 1.7E-06 | 1.4E-05 | 1.7E-06 | 1.7E-06 |
| vs. WO | 1.7E-06 | 2.0E-05 | 1.6E-01 | 1.6E-03 | 8.6E-01 | 5.7E-01 | 3.1E-04 | 7.2E-02 | 4.1E-02 | 1.2E-02 | 2.2E-05 | 2.2E-02 |
| vs. NRBO | 1.7E-06 | 9.3E-06 | 1.7E-06 | 1.7E-06 | 1.7E-06 | 4.7E-03 | 1.7E-06 | 3.2E-06 | 1.7E-06 | 1.1E-04 | 1.7E-06 | 8.2E-03 |

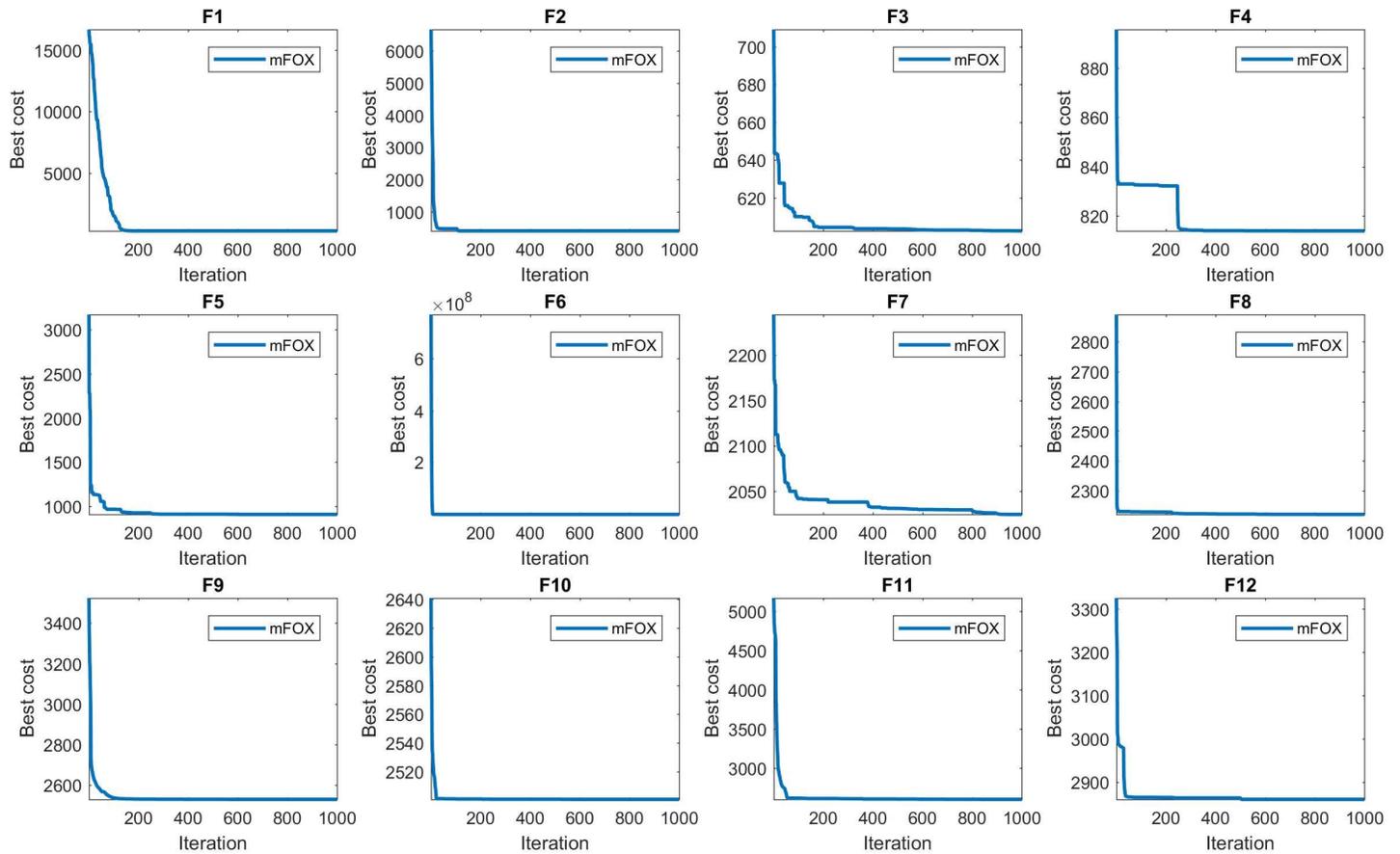

**Figure 7.** Convergence behavior analysis of mFOX for the CEC2022 benchmark test problems.

## 3.4. Real-world applications

This section evaluates the performance of the mFOX algorithm on real-world constrained engineering optimization problems from the CEC2020 benchmark [53]. Four problems were selected: tension/compression spring, pressure vessel design, gas transmission compressor design, and hydrostatic thrust-bearing design. The results of mFOX were compared against twelve other algorithms, as listed in Table 1. A static penalty function approach is employed to manage the constraints of each problem. Each algorithm is executed with a population size of 30 and subjected to 30,000 function evaluations, ensuring a comprehensive comparison. Performance metrics, including standard deviation, average, minimum, and maximum values, are analyzed after 30 independent runs for each problem. A statistical analysis of the results demonstrates how effectively the mFOX algorithm handles constraints during the optimization process. Table 12 also outlines the details of the selected problems.

**Table 12:** Details of all selected real-world constrained functions from CEC2020 [53].

| Name of the problem | dim | Inequality constraints (g) | Equality constraints (h) | $f_{min}$ |
|---|---|---|---|---|
| Tension/compression spring design (case 1) | 3 | 3 | 0 | 0.012665232788 |
| Pressure vessel design | 4 | 4 | 0 | 5885.3327736 |
| Gas Transmission Compressor Design | 4 | 1 | 0 | 2,964,895.4173 |
| Hydro-static thrust bearing design problem | 4 | 7 | 0 | 1625.4428092 |

### 3.4.1. Tension/compression spring design problem

The primary goal of this problem, depicted in Figure 8.a, is to achieve weight reduction in a spring. The optimization involves three design variables: mean coil diameter ($D$), wire diameter ($d$), and the number of active coils ($N$). The solution must satisfy four critical constraints: deflection ($g_1$), shear stress ($g_2$), surge frequency ($g_3$), and outer diameter limit ($g_4$) [54]. These constraints are articulated mathematically in Eq. (18).

Table 13 presents a comparison of the best solutions achieved by the mFOX algorithm and its competitors, along with their statistical results. Notably, mFOX achieves the best average solution among the comparable algorithms, highlighting the algorithm's effectiveness in addressing the tension/compression spring problem. Figure 8.b illustrates the convergence behavior of the mFOX algorithm as it optimizes the tension/compression spring design. The curve highlights the algorithm's progressive approach toward achieving the optimal solution, demonstrating its efficiency and effectiveness in this design problem. This visual evidence underscores the mFOX algorithm's capability in solving complex engineering optimization challenges.

$$
\begin{aligned}
&Consider:\quad \bar{x} = [x_1\, x_2\, x_3] = [d\ D\ N], \\
&Minimize:\quad f(\bar{x}) = (x_3 + 2)x_2 x_1^2, \\
&Subject\,to: \\
&\quad g_1(\bar{x}) = 1 - \frac{x_2^3 x_3}{71785 x_1^4} \leq 0, \\
&\quad g_2(\bar{x}) = \frac{4x_2^2 - x_1 x_2}{12566(x_1^3 x_2 - x_1^4)} + \frac{1}{5108 x_1^2} - 1 \leq 0, \\
&\quad g_3(\bar{x}) = 1 - \frac{140.45 x_1}{x_2^2 x_3} \leq 0,
\end{aligned}
\qquad (18)
$$

$$g_4(\bar{x}) = \frac{x_1 + x_2}{1.5} - 1 \leq 0.$$

Variable range: $0.05 \leq x1 \leq 2.00, 0.25 \leq x2 \leq 1.30, 2.00 \leq x3 \leq 15.0$

**Table 13:** Experimental results of Tension/compression spring design (*SIs* are statistical indicators and *DVs* are design variables).

|     |       | mFOX | FOX | WOA | TSA | ChOA | GWO | AZOA | HO | ROA | WO | NRBO |
|-----|-------|------|-----|-----|-----|------|-----|------|-----|-----|-----|-----|
| SIs | AVG   | **0.01269** | 0.012906 | 0.013668 | 0.012899 | 0.013284 | 0.012743 | 0.014401 | 0.012969 | 7.6E+12 | 0.013646 | 0.013102 |
|     | Best  | 0.012666 | 0.012674 | 0.012668 | 0.012684 | 0.012765 | 0.012677 | 0.012681 | 0.012688 | 0.013193 | 0.012665 | 0.012666 |
|     | Worst | 0.012766 | 0.014363 | 0.016947 | 0.013414 | 0.015535 | 0.01305 | 0.017773 | 0.014443 | 2.28E+14 | 0.017773 | 0.015296 |
|     | STD   | 2.64E-05 | 0.00038 | 0.001219 | 0.000187 | 0.000679 | 7.55E-05 | 0.002075 | 0.000373 | 4.09E+13 | 0.001608 | 0.000593 |
|     | RANK  | 1 | 4 | 9 | 3 | 7 | 2 | 10 | 5 | 11 | 8 | 6 |
| DVs | $X_1$ | 0.051686 | 0.051792 | 0.051271 | 0.051713 | 0.05 | 0.051499 | 0.050765 | 0.05266 | 0.05 | 0.051574 | 0.051889 |
|     | $X_2$ | 0.356639 | 0.359121 | 0.346751 | 0.357063 | 0.317381 | 0.352148 | 0.334902 | 0.380521 | 0.310417 | 0.353964 | 0.361554 |
|     | $X_3$ | 11.29403 | 11.15625 | 11.89813 | 11.28406 | 14.08824 | 11.5733 | 12.69268 | 10.02376 | 15 | 11.45225 | 11.01099 |

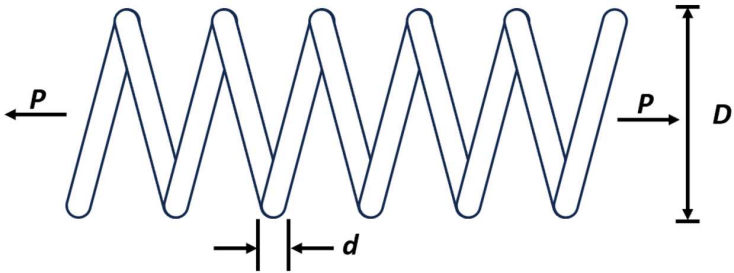
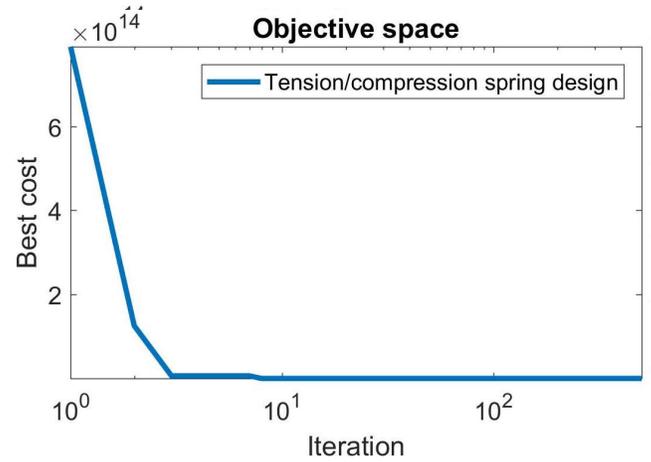

**a.** Schematic design  **b.** Function values over iterations

**Figure 8.** Tension/compression spring design problem.

### 3.4.2. Pressure Vessel Design Problem

This section addresses the optimization of cost in the construction of cylindrical pressure vessels, with a focus on meeting specified pressure requirements. The goal is to achieve a cost-effective design while maintaining adherence to critical safety and performance standards [55]. The primary design variables include the length of the cylindrical section ($L$), the inner and head radii ($R$), as well as the thickness of the head ($T_h$) and the shell ($T_s$). These parameters are essential for effective pressure vessel design. Additionally, the design must conform to constraints concerning buckling load, end deflection, shear stress, and bending stress. The nature of this optimization problem is illustrated in Figure 9.a, with the fitness function expressed in Eq. (19).

The optimization and statistical results for pressure vessel design using the mFOX and competing algorithms are summarized in Table 14. The mFOX achieved the best solution with design variable values of (0.778175,

0.384652, 40.31996, 199.9994) and an objective function value of 5885.432. These results demonstrate that the mFOX outperforms other algorithms in effectively solving pressure vessel design problems. Figure 9.b presents the convergence curve of the mFOX algorithm, depicting its search process toward identifying the optimal solution.

$$\text{Consider:} \quad \bar{x} = [x_1 x_2 x_3 x_4] = [T_s \ T_h \ R \ L],$$

$$\text{Minimize:} \quad f(\bar{x}) = 0.6224 x_1 x_3 x_4 + 1.7781 x_2 x_3^2 + 3.1661 x_1^2 x_4 + 19.84 x_1^2 x_3,$$

$$\text{Subject to:} \quad g_1(\bar{x}) = -x_2 + 0.00954 x_3 \leq 0,$$

$$g_2(\bar{x}) = -x_1 + 0.0193 x_3 \leq 0, \quad (19)$$

$$g_3(\bar{x}) = x_4 - 240 \leq 0.$$

$$g_4(\bar{x}) = -\pi x_3^2 x_4 - \frac{4}{3} \pi x_3^3 + 1296000 \leq 0,$$

$$\text{Variable range:} \quad 0 \leq x1 \leq 100, 0 \leq x2 \leq 100, 10 \leq x3 \leq 200, 10 \leq x4 \leq 200.$$

**Table 14:** Experimental results of Pressure Vessel Design (*SIs* are statistical indicators and *DVs* are design variables).

| | | mFOX | FOX | WOA | TSA | ChOA | GWO | AZOA | HO | ROA | WO | NRBO |
|---|---|---|---|---|---|---|---|---|---|---|---|---|
| SIs | AVG | **6046.037** | 72465.79 | 9208.661 | 6279.046 | 7839.883 | 6185.878 | 6757.608 | 6568.666 | 35895.59 | 6526.846 | 6577.73 |
| | Best | 5885.432 | 5979.305 | 6279.271 | 5886.724 | 6972.326 | 5895.286 | 5885.343 | 6003.558 | 6941.093 | 5885.931 | 5885.528 |
| | Worst | 6801.196 | 262941.8 | 23888.47 | 7385.415 | 8337.662 | 7295.787 | 8176.93 | 7326.507 | 108111.5 | 7319.28 | 7319.025 |
| | STD | 228.8301 | 81982.66 | 3260.386 | 514.6217 | 301.7565 | 453.4331 | 590.6693 | 361.787 | 30880.58 | 533.2597 | 482.7469 |
| | RANK | 1 | 11 | 9 | 3 | 8 | 2 | 7 | 5 | 10 | 4 | 6 |
| DVs | $T_s$ | 0.778175 | 0.828306 | 0.901937 | 0.77828 | 1.136172 | 0.77952 | 0.778169 | 0.838328 | 1.093505 | 0.778511 | 0.778169 |
| | $T_h$ | 0.384652 | 0.409591 | 0.481508 | 0.384756 | 0.539033 | 0.387621 | 0.384653 | 0.41472 | 0.579322 | 0.384821 | 0.384717 |
| | $R$ | 40.31996 | 42.91616 | 46.65208 | 40.3213 | 55.6182 | 40.38744 | 40.31962 | 43.39487 | 56.21628 | 40.33732 | 40.31962 |
| | $L$ | 199.9994 | 166.8412 | 127.3426 | 200 | 59.49209 | 199.0887 | 200 | 161.2488 | 55.93956 | 199.7538 | 200 |

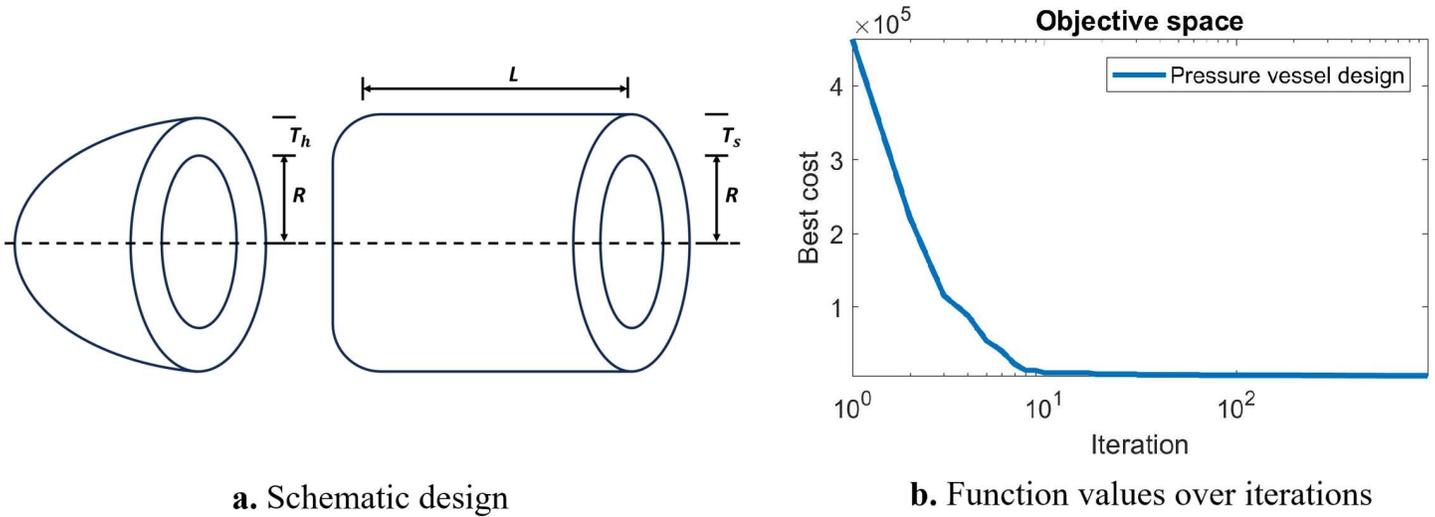

**a.** Schematic design  **b.** Function values over iterations

**Figure 9.** Pressure vessel design problem.

### 3.4.3. Gas Transmission Compressor Design

The gas transmission compressor design problem focuses on optimizing the compressor's performance with four design variables and one constraint [53]. The objective is to minimize the design function, as formulated in Eq. (20) and depicted in Figure 10.a [56].

Table 15 presents the optimization and statistical results obtained from applying the mFOX algorithm alongside various competing algorithms for the design optimization of gas transmission compressors. The findings indicate that the proposed mFOX approach achieved the optimal design, as reflected in the results obtained. The identified design variables were (50, 1.178568, 24.51979, 0.389022), leading to an objective function value of 2,964,901.40897659. Furthermore, the simulation results highlight that mFOX significantly outperforms other algorithms in effectively addressing the gas transmission compressor design problem. Figure 10.b presents the convergence curve of the mFOX algorithm, showcasing its ability to achieve the optimal design. This performance highlights the algorithm's potential to improve the efficiency of gas transmission compressor design.

$$
\begin{aligned}
Minimize: \quad & f(\bar{x}) = 8.61 \times 10^5 x_1^{\frac{1}{2}} x_2 x_3^{-\frac{2}{3}} x_4^{-\frac{1}{2}} + 3.69 \times 10^4 x_3 + 7.72 \times 10^8 x_1^{-1} x_2^{0.219} - 765.43 \times 10^6 x_1^{-1}, \\
Subject\ to: \quad & x_4 x_2^{-2} + x_2^{-2} - 1 \leq 0, \\
Variable\ range: \quad & 20 \leq x_1 \leq 50, 1 \leq x_2 \leq 10, 20 \leq x_3 \leq 50, 0.1 \leq x_4 \leq 60,
\end{aligned}
\tag{20}
$$

**Table 15:** Experimental results of gas transmission compressor design (*SIs* are statistical indicators and *DVs* are design variables).

|     |       | mFOX        | FOX      | WOA      | TSA      | ChOA     | GWO      | AZOA     | HO       | ROA      | WO       | NRBO     |
|-----|-------|-------------|----------|----------|----------|----------|----------|----------|----------|----------|----------|----------|
| SIs | AVG   | **2965200** | 3096072  | 3004237  | 2966368  | 2974580  | 2965344  | 2990420  | 2978842  | 3145356  | 2998158  | 2970268  |
|     | Best  | 2964901     | 3077549  | 2965003  | 2965228  | 2965806  | 2964923  | 2964895  | 2965507  | 3063587  | 2964944  | 2964896  |
|     | Worst | 2966473     | 3105291  | 3187742  | 2969360  | 2984493  | 2966206  | 3358535  | 3077032  | 3187779  | 3096175  | 2997693  |
|     | STD   | 395.7626    | 6722.775 | 47745.82 | 949.2433 | 5307.531 | 316.3391 | 76046.62 | 20447.06 | 43907.14 | 42947.09 | 9262.738 |

| | RANK | 1 | 10 | 9 | 3 | 5 | 2 | 7 | 6 | 11 | 8 | 4 |
|---|---|---|---|---|---|---|---|---|---|---|---|---|
| DVs | $X_1$ | 50 | 21.82094 | 49.85936 | 50 | 50 | 50 | 50 | 49.91498 | 22.8507 | 49.9011 | 49.99817 |
| | $X_2$ | 1.178568 | 1.083744 | 1.179569 | 1.173207 | 1.181527 | 1.179101 | 1.178284 | 1.177194 | 1.078491 | 1.177184 | 1.178315 |
| | $X_3$ | 24.51979 | 20.70732 | 24.38228 | 24.68591 | 24.23617 | 24.62454 | 24.59258 | 25.30174 | 23.25849 | 24.69542 | 24.58571 |
| | $X_4$ | 0.389022 | 0.1745 | 0.391382 | 0.376398 | 0.395593 | 0.39027 | 0.388353 | 0.385785 | 0.161856 | 0.385762 | 0.388426 |

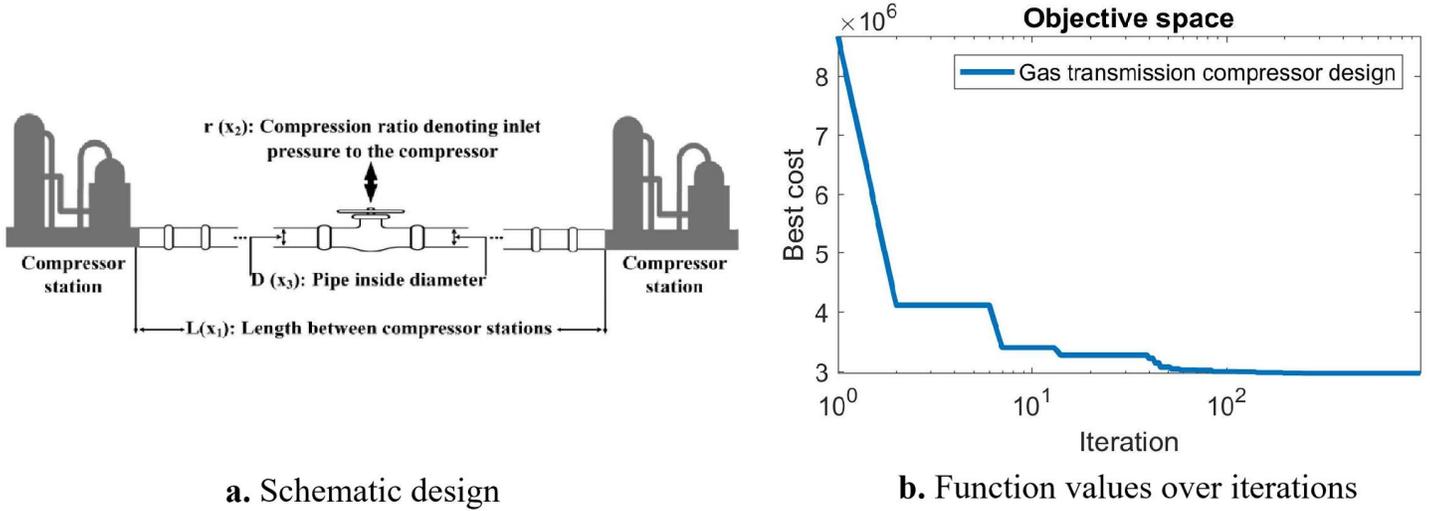

a. Schematic design  b. Function values over iterations

**Figure 10.** Gas transmission compressor design.

### 3.4.4. Hydro-Static Thrust-Bearing Design Problem

The hydro-static thrust-bearing design problem focuses on optimizing bearing power losses through four key design variables: oil viscosity, bearing radius, flow rate, and groove radius [53]. The problem is subject to seven nonlinear constraints, represented mathematically in Eq. (21):

Table 16 presents the results of the mFOX algorithm in comparison with other competing algorithms for optimizing the design of hydro-static thrust-bearing. The findings illustrate that the mFOX algorithm effectively identifies the optimal design solution for the hydro-static thrust-bearing design challenge. Specifically, the optimized model variables are recorded as (6.009451, 5.446845, 5.37E-06, 2.298136), which correspond to an objective function value of 1637.58078556918. The simulation outcomes indicate that the mFOX algorithm outperforms its counterparts in addressing the hydro-static thrust-bearing design problem, consistently yielding superior results. The effectiveness of the mFOX algorithm is further demonstrated in Figure 11, which presents its convergence curve. This curve visually captures the algorithm's progression toward the optimal design, highlighting its strong performance in tackling the optimization task.

$$\begin{aligned}
\text{Minimize:} \quad & f(\bar{x}) = \frac{QP_0}{0.7} + E_f, \\
\text{Subject to:} \quad & g_1(\bar{x}) = 1000 - P_0 \leq 0, \\
& g_2(\bar{x}) = W - 101000 \leq 0, \\
& g_3(\bar{x}) = 5000 - \frac{W}{\pi(R^2 - R_0^2)} \leq 0,
\end{aligned} \quad (21)$$

$$g_4(\bar{x}) = 50 - P_0 \le 0,$$

$$g_5(\bar{x}) = 0.001 - \frac{0.0307}{386.4 P_0}\left(\frac{Q}{2\pi Rh}\right) \le 0,$$

$$g_6(\bar{x}) = R - R_0 \le 0,$$

$$g_7(\bar{x}) = h - 0.001 \le 0,$$

Where:
$$W = \frac{\pi P_0}{2}\frac{R^2 - R_0^2}{\ln\left(\frac{R}{R_0}\right)},$$

$$P_0 = \frac{6\mu Q}{\pi h^3}\ln\left(\frac{R}{R_0}\right),$$

$$E_f = 9336 Q \times 0.0307 \times 0.5 \Delta T, \Delta T = 2(10^P - 559.7),$$

$$P = \frac{\log_{10}\log_{10}(8.122 \times 10^6 \mu + 0.8) + 3.55}{10.04},$$

$$h = \left(\frac{2\pi \times 750}{60}\right)^2 \frac{2\pi\mu}{E_f}\left(\frac{R^4}{4} - \frac{R_0^4}{4}\right),$$

Variable range:   $1 \le R \le 16, 1 \le R_0 \le 16, 1 \times 10^{-6} \le \mu \le 16 \times 10^{-6}, 1 \le Q \le 16,.$

**Table 16:** Experimental results of hydro-static thrust-bearing design problem (*SIs* are statistical indicators and *DVs* are design variables).

|     |       | mFOX | FOX | WOA | TSA | ChOA | GWO | AZOA | HO | ROA | WO | NRBO |
|-----|-------|------|-----|-----|-----|------|-----|------|----|-----|----|------|
| SIs | AVG   | **1826.708** | 3.76E+21 | 1.16E+08 | 9.36E+08 | 1.49E+08 | 2065.061 | 2430.889 | 2967.059 | 9.04E+08 | 47026657 | 2597.126 |
|     | Best  | 1637.581 | 4545.7 | 1908.724 | 1720.252 | 2474.382 | 1744.538 | 1711.452 | 2128.561 | 2722.407 | 1888.076 | 1667.747 |
|     | Worst | 2211.714 | 4.86E+22 | 1.27E+09 | 2.2E+10 | 4.46E+09 | 2693.679 | 3467.989 | 4035.802 | 3.09E+09 | 1.41E+09 | 4941.096 |
|     | STD   | 123.5381 | 1.1E+22 | 3.5E+08 | 3.99E+09 | 8.01E+08 | 225.4277 | 468.5164 | 529.3406 | 9.72E+08 | 2.53E+08 | 655.4117 |
|     | RANK  | 1 | 11 | 7 | 10 | 8 | 2 | 3 | 5 | 9 | 6 | 4 |
| DVs | $X_1$ | 6.009451 | 8.582246 | 6.532993 | 6.0081 | 7.445659 | 6.041103 | 6.166909 | 6.467478 | 8.168915 | 6.120168 | 6.006959 |
|     | $X_2$ | 5.446845 | 7.910116 | 5.964619 | 5.438444 | 6.969985 | 5.478047 | 5.621428 | 5.949097 | 7.751809 | 5.569694 | 5.445521 |
|     | $X_3$ | 5.37E-06 | 7.09E-06 | 5.37E-06 | 6E-06 | 7.42E-06 | 6.15E-06 | 5.55E-06 | 7.94E-06 | 7.61E-06 | 7.17E-06 | 5.67E-06 |
|     | $X_4$ | 2.298136 | 10.84079 | 2.756752 | 2.862963 | 6.934406 | 3.025284 | 2.547976 | 6.667484 | 8.526007 | 4.443937 | 2.537141 |

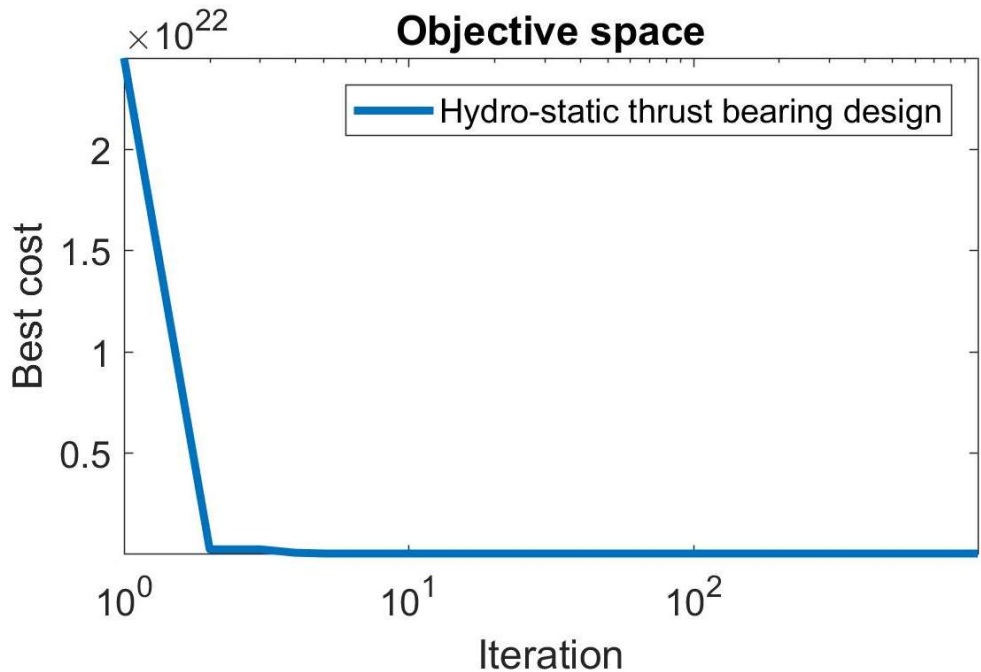

**Figure 11.** Hydro-static thrust-bearing design.

## 4. Conclusion and Future Work

In this work, a modification of the FOX optimizer is introduced as a novel algorithm called mFOX. The key enhancement focuses on improving the exploration phase to achieve a balanced trade-off between exploration and exploitation. The mFOX algorithm has been evaluated on 45 benchmark functions, including 23 classical benchmark functions, 10 CEC2019 functions, and 12 CEC2022 benchmark test functions. Additionally, four real-world engineering problems were used to assess the validity of mFOX: gas transmission compressor design, hydrostatic thrust-bearing design, tension/compression spring, and pressure vessel design. The results demonstrate the efficiency and superiority of mFOX compared to WOA, TSA, ChOA, FDO, GWO, DA, AZOA, HO, ROA, WO, NRBO, and FOX.

Several research directions can be proposed for future work. One possibility is developing mFOX variants, incorporating discretization techniques and multi-objective optimization strategies. Additionally, mFOX could be applied to address complex combinatorial optimization problems, including the Vehicle Routing Problem, dimensionality reduction, network design, and image segmentation. Future research could also explore hybridizing mFOX with other heuristic and metaheuristic algorithms to enhance its performance. Moreover, mFOX could also be utilized for feature selection.

**Acknowledgments:** The authors wish to thank Charmo University.

**Funding:** This research received no external funding.

**Code availability:** The source code of the mFOX algorithm is available upon reasonable request from the corresponding author.

**Data availability**: All data are incorporated into the article.

**Conflicts of Interest:** The authors declare that they have no conflict of interest.

# References

[1] M. Dehghani, E. Trojovská, P. Trojovský, and O. P. Malik, "OOBO: A New Metaheuristic Algorithm for Solving Optimization Problems," *Biomimetics*, vol. 8, no. 6, 2023, doi: 10.3390/biomimetics8060468.

[2] V. Tomar, M. Bansal, and P. Singh, "Metaheuristic Algorithms for Optimization: A Brief Review," *Engineering Proceedings*, vol. 59, no. 1, 2023, doi: 10.3390/engproc2023059238.

[3] E. H. Houssein, D. Oliva, E. Çelik, M. M. Emam, and R. M. Ghoniem, "Boosted sooty tern optimization algorithm for global optimization and feature selection," *Expert Syst Appl*, vol. 213, p. 119015, 2023, doi: https://doi.org/10.1016/j.eswa.2022.119015.

[4] V. Rajinikanth and N. Razmjooy, "A Comprehensive Survey of Meta-heuristic Algorithms," in *Metaheuristics and Optimization in Computer and Electrical Engineering: Volume 2: Hybrid and Improved Algorithms*, N. Razmjooy, N. Ghadimi, and V. Rajinikanth, Eds., Cham: Springer International Publishing, 2023, pp. 1–39. doi: 10.1007/978-3-031-42685-8_1.

[5] B. Singh and M. Murugaiah, "Bio-inspired Computing and Associated Algorithms," in *High Performance Computing in Biomimetics: Modeling, Architecture and Applications*, K. A. Ahmad, N. A. W. A. Hamid, M. Jawaid, T. Khan, and B. Singh, Eds., Singapore: Springer Nature Singapore, 2024, pp. 47–87. doi: 10.1007/978-981-97-1017-1_3.

[6] D. Tuličić, M. Horvat, and S. Lovrenčić, "Towards Swarm Intelligence Ontology for Formal Description of Metaheuristics Algorithms," in *2023 International Conference on Computing, Networking, Telecommunications & Engineering Sciences Applications (CoNTESA)*, 2023, pp. 43–47. doi: 10.1109/CoNTESA61248.2023.10384887.

[7] A. Tzanetos and G. Dounias, "Nature inspired optimization algorithms or simply variations of metaheuristics?," *Artif Intell Rev*, vol. 54, no. 3, pp. 1841–1862, 2021, doi: 10.1007/s10462-020-09893-8.

[8] A. Yaqoob, N. K. Verma, and R. M. Aziz, "Metaheuristic Algorithms and Their Applications in Different Fields," in *Metaheuristics for Machine Learning*, 2024, pp. 1–35. doi: https://doi.org/10.1002/9781394233953.ch1.

[9] J. H. Holland, "Genetic Algorithms," *Sci Am*, vol. 267, no. 1, pp. 66–73, 1992, [Online]. Available: http://www.jstor.org/stable/24939139

[10] R. Storn and K. Price, "Differential Evolution – A Simple and Efficient Heuristic for global Optimization over Continuous Spaces," *Journal of Global Optimization*, vol. 11, no. 4, pp. 341–359, 1997, doi: 10.1023/A:1008202821328.

[11] H.-G. Beyer and H.-P. Schwefel, "Evolution strategies – A comprehensive introduction," *Nat Comput*, vol. 1, no. 1, pp. 3–52, 2002, doi: 10.1023/A:1015059928466.

[12] J. Tang, G. Liu, and Q. Pan, "A Review on Representative Swarm Intelligence Algorithms for Solving Optimization Problems: Applications and Trends," *IEEE/CAA Journal of Automatica Sinica*, vol. 8, no. 10, pp. 1627–1643, 2021, doi: 10.1109/JAS.2021.1004129.

[13] J. Kennedy and R. Eberhart, "Particle swarm optimization," in *Proceedings of ICNN'95 - International Conference on Neural Networks*, 1995, pp. 1942–1948 vol.4. doi: 10.1109/ICNN.1995.488968.

[14] M. Dorigo, M. Birattari, and T. Stutzle, "Ant colony optimization," *IEEE Comput Intell Mag*, vol. 1, no. 4, pp. 28–39, 2006, doi: 10.1109/MCI.2006.329691.

[15] S. Mirjalili, S. M. Mirjalili, and A. Lewis, "Grey Wolf Optimizer," *Advances in Engineering Software*, vol. 69, pp. 46–61, 2014, doi: https://doi.org/10.1016/j.advengsoft.2013.12.007.

[16]   A. Faramarzi, M. Heidarinejad, S. Mirjalili, and A. H. Gandomi, "Marine Predators Algorithm: A nature-inspired metaheuristic," *Expert Syst Appl*, vol. 152, p. 113377, 2020, doi: https://doi.org/10.1016/j.eswa.2020.113377.

[17]   S. Kaur, L. K. Awasthi, A. L. Sangal, and G. Dhiman, "Tunicate Swarm Algorithm: A new bio-inspired based metaheuristic paradigm for global optimization," *Eng Appl Artif Intell*, vol. 90, p. 103541, 2020, doi: https://doi.org/10.1016/j.engappai.2020.103541.

[18]   N. Chopra and M. Mohsin Ansari, "Golden jackal optimization: A novel nature-inspired optimizer for engineering applications," *Expert Syst Appl*, vol. 198, p. 116924, 2022, doi: https://doi.org/10.1016/j.eswa.2022.116924.

[19]   S. Mirjalili and A. Lewis, "The Whale Optimization Algorithm," *Advances in Engineering Software*, vol. 95, pp. 51–67, 2016, doi: https://doi.org/10.1016/j.advengsoft.2016.01.008.

[20]   M. Khishe and M. R. Mosavi, "Chimp optimization algorithm," *Expert Syst Appl*, vol. 149, p. 113338, 2020, doi: https://doi.org/10.1016/j.eswa.2020.113338.

[21]   J. M. Abdullah and T. Ahmed, "Fitness Dependent Optimizer: Inspired by the Bee Swarming Reproductive Process," *IEEE Access*, vol. 7, pp. 43473–43486, 2019, doi: 10.1109/ACCESS.2019.2907012.

[22]   S. Mirjalili, "Dragonfly algorithm: a new meta-heuristic optimization technique for solving single-objective, discrete, and multi-objective problems," *Neural Comput Appl*, vol. 27, no. 4, pp. 1053–1073, 2016, doi: 10.1007/s00521-015-1920-1.

[23]   S. Mohapatra and P. Mohapatra, "American zebra optimization algorithm for global optimization problems," *Sci Rep*, vol. 13, no. 1, p. 5211, 2023, doi: 10.1038/s41598-023-31876-2.

[24]   M. H. Amiri, N. Mehrabi Hashjin, M. Montazeri, S. Mirjalili, and N. Khodadadi, "Hippopotamus optimization algorithm: a novel nature-inspired optimization algorithm," *Sci Rep*, vol. 14, no. 1, p. 5032, 2024, doi: 10.1038/s41598-024-54910-3.

[25]   H. Jia, X. Peng, and C. Lang, "Remora optimization algorithm," *Expert Syst Appl*, vol. 185, p. 115665, 2021, doi: https://doi.org/10.1016/j.eswa.2021.115665.

[26]   R. Sowmya, M. Premkumar, and P. Jangir, "Newton-Raphson-based optimizer: A new population-based metaheuristic algorithm for continuous optimization problems," *Eng Appl Artif Intell*, vol. 128, p. 107532, 2024, doi: https://doi.org/10.1016/j.engappai.2023.107532.

[27]   M. Han, Z. Du, K. F. Yuen, H. Zhu, Y. Li, and Q. Yuan, "Walrus optimizer: A novel nature-inspired metaheuristic algorithm," *Expert Syst Appl*, vol. 239, p. 122413, 2024, doi: https://doi.org/10.1016/j.eswa.2023.122413.

[28]   M. Dehghani, E. Trojovská, and P. Trojovský, "A new human-based metaheuristic algorithm for solving optimization problems on the base of simulation of driving training process," *Sci Rep*, vol. 12, no. 1, p. 9924, 2022, doi: 10.1038/s41598-022-14225-7.

[29]   M. Hubalovska and S. Major, "A New Human-Based Metaheuristic Algorithm for Solving Optimization Problems Based on Technical and Vocational Education and Training," *Biomimetics*, vol. 8, no. 6, 2023, doi: 10.3390/biomimetics8060508.

[30]   M. Dehghani *et al.*, "A New 'Doctor and Patient' Optimization Algorithm: An Application to Energy Commitment Problem," *Applied Sciences*, vol. 10, no. 17, 2020, doi: 10.3390/app10175791.

[31] S. O. Oladejo, S. O. Ekwe, and S. Mirjalili, "The Hiking Optimization Algorithm: A novel human-based metaheuristic approach," *Knowl Based Syst*, vol. 296, p. 111880, 2024, doi: https://doi.org/10.1016/j.knosys.2024.111880.

[32] M. Abdel-Basset, R. Mohamed, S. A. A. Azeem, M. Jameel, and M. Abouhawwash, "Kepler optimization algorithm: A new metaheuristic algorithm inspired by Kepler's laws of planetary motion," *Knowl Based Syst*, vol. 268, p. 110454, 2023, doi: https://doi.org/10.1016/j.knosys.2023.110454.

[33] E. Rashedi, H. Nezamabadi-pour, and S. Saryazdi, "GSA: A Gravitational Search Algorithm," *Inf Sci (N Y)*, vol. 179, no. 13, pp. 2232–2248, 2009, doi: https://doi.org/10.1016/j.ins.2009.03.004.

[34] A. Sadollah, H. Eskandar, H. M. Lee, D. G. Yoo, and J. H. Kim, "Water cycle algorithm: A detailed standard code," *SoftwareX*, vol. 5, pp. 37–43, 2016, doi: https://doi.org/10.1016/j.softx.2016.03.001.

[35] A. Faramarzi, M. Heidarinejad, B. Stephens, and S. Mirjalili, "Equilibrium optimizer: A novel optimization algorithm," *Knowl Based Syst*, vol. 191, p. 105190, 2020, doi: https://doi.org/10.1016/j.knosys.2019.105190.

[36] M. Dehghani *et al.*, "A Spring Search Algorithm Applied to Engineering Optimization Problems," *Applied Sciences*, vol. 10, no. 18, 2020, doi: 10.3390/app10186173.

[37] H. Mohammed and T. Rashid, "FOX: a FOX-inspired optimization algorithm," *Applied Intelligence*, vol. 53, no. 1, pp. 1030–1050, 2023, doi: 10.1007/s10489-022-03533-0.

[38] M. S. Painter *et al.*, "Use of bio-loggers to characterize red fox behavior with implications for studies of magnetic alignment responses in free-roaming animals," *Animal Biotelemetry*, vol. 4, no. 1, p. 20, 2016, doi: 10.1186/s40317-016-0113-8.

[39] J. Červený, S. Begall, P. Koubek, P. Nováková, and H. Burda, "Directional preference may enhance hunting accuracy in foraging foxes," *Biol Lett*, vol. 7, no. 3, pp. 355–357, 2011, doi: https://doi.org/10.1098/rsbl.2010.1145.

[40] K. Meidani, A. Hemmasian, S. Mirjalili, and A. Barati Farimani, "Adaptive grey wolf optimizer," *Neural Comput Appl*, vol. 34, no. 10, pp. 7711–7731, 2022, doi: 10.1007/s00521-021-06885-9.

[41] F. K. Karim, D. S. Khafaga, M. M. Eid, S. K. Towfek, and H. K. Alkahtani, "A Novel Bio-Inspired Optimization Algorithm Design for Wind Power Engineering Applications Time-Series Forecasting," *Biomimetics*, vol. 8, no. 3, 2023, doi: 10.3390/biomimetics8030321.

[42] H. R. Tizhoosh, "Opposition-Based Learning: A New Scheme for Machine Intelligence," in *International Conference on Computational Intelligence for Modelling, Control and Automation and International Conference on Intelligent Agents, Web Technologies and Internet Commerce (CIMCA-IAWTIC'06)*, 2005, pp. 695–701. doi: 10.1109/CIMCA.2005.1631345.

[43] E. H. Houssein *et al.*, "An improved opposition-based marine predators algorithm for global optimization and multilevel thresholding image segmentation," *Knowl Based Syst*, vol. 229, p. 107348, 2021, doi: https://doi.org/10.1016/j.knosys.2021.107348.

[44] D. Izci, S. Ekinci, E. Eker, and M. Kayri, "Augmented hunger games search algorithm using logarithmic spiral opposition-based learning for function optimization and controller design," *Journal of King Saud University - Engineering Sciences*, vol. 36, no. 5, pp. 330–338, 2024, doi: https://doi.org/10.1016/j.jksues.2022.03.001.

[45] M. Barhoush, B. H. Abed-alguni, and N. E. A. Al-qudah, "Improved discrete salp swarm algorithm using exploration and exploitation techniques for feature selection in intrusion detection systems," *J Supercomput*, vol. 79, no. 18, pp. 21265–21309, 2023, doi: 10.1007/s11227-023-05444-4.

[46] V. H. S. Pham, N. T. Nguyen Dang, and V. N. Nguyen, "Enhancing Global Optimization through the Integration of Multiverse Optimizer with Opposition-Based Learning," *Applied Computational Intelligence and Soft Computing*, vol. 2024, no. 1, p. 6661599, 2024, doi: 10.1155/2024/6661599.

[47] M. Tubishat, M. A. M. Abushariah, N. Idris, and I. Aljarah, "Improved whale optimization algorithm for feature selection in Arabic sentiment analysis," *Applied Intelligence*, vol. 49, no. 5, pp. 1688–1707, 2019, doi: 10.1007/s10489-018-1334-8.

[48] N. Alamir, S. Kamel, M. H. Hassan, and S. M. Abdelkader, "An effective quantum artificial rabbits optimizer for energy management in microgrid considering demand response," *Soft comput*, vol. 27, no. 21, pp. 15741–15768, 2023, doi: 10.1007/s00500-023-08814-5.

[49] A. A. H. Amin, A. M. Aladdin, D. O. Hasan, S. R. Mohammed-Taha, and T. A. Rashid, "Enhancing Algorithm Selection through Comprehensive Performance Evaluation: Statistical Analysis of Stochastic Algorithms," *Computation*, vol. 11, no. 11, 2023, doi: 10.3390/computation11110231.

[50] P. N. Suganthan *et al.*, "Problem definitions and evaluation criteria for the CEC 2005 special session on real-parameter optimization," *KanGAL report*, vol. 2005005, no. 2005, p. 2005, 2005.

[51] K. V Price, N. H. Awad, M. Z. Ali, and P. N. Suganthan, "Problem definitions and evaluation criteria for the 100-digit challenge special session and competition on single objective numerical optimization," in *Technical report*, Nanyang Technological University Singapore, 2018.

[52] D. Yazdani *et al.*, "IEEE CEC 2022 competition on dynamic optimization problems generated by generalized moving peaks benchmark," *arXiv preprint arXiv:2106.06174*, 2021.

[53] A. Kumar, G. Wu, M. Z. Ali, R. Mallipeddi, P. N. Suganthan, and S. Das, "A test-suite of non-convex constrained optimization problems from the real-world and some baseline results," *Swarm Evol Comput*, vol. 56, p. 100693, 2020, doi: https://doi.org/10.1016/j.swevo.2020.100693.

[54] R. M. Rizk-Allah and E. Elsodany, "An improved rough set strategy-based sine cosine algorithm for engineering optimization problems," *Soft comput*, pp. 1–22, 2023.

[55] D. Zou, H. Liu, L. Gao, and S. Li, "A novel modified differential evolution algorithm for constrained optimization problems," *Computers & Mathematics with Applications*, vol. 61, no. 6, pp. 1608–1623, 2011, doi: https://doi.org/10.1016/j.camwa.2011.01.029.

[56] M. H. Nadimi-Shahraki, A. Fatahi, H. Zamani, S. Mirjalili, and L. Abualigah, "An improved moth-flame optimization algorithm with adaptation mechanism to solve numerical and mechanical engineering problems," *Entropy*, vol. 23, no. 12, p. 1637, 2021, doi: https://doi.org/10.3390/e23121637.